\documentclass[final]{cvpr}

\usepackage[pagebackref=true,breaklinks=true,colorlinks,bookmarks=false]{hyperref}

\def\cvprPaperID{11437} 
\def\confYear{CVPR 2021}
\usepackage{times}
\usepackage{epsfig}
\usepackage{graphicx}
\usepackage{amsmath}
\usepackage{amssymb}

\usepackage{amsmath,amsfonts,bm}

\def\eqref#1{equation~\ref{#1}}

\def\1{\bm{1}}

\DeclareMathAlphabet{\mathsfit}{\encodingdefault}{\sfdefault}{m}{sl}
\SetMathAlphabet{\mathsfit}{bold}{\encodingdefault}{\sfdefault}{bx}{n}

\def\emB{{B}}

\DeclareMathOperator*{\argmax}{arg\,max}

\usepackage{arydshln}

\newcommand{\specialcell}[2][c]{%
	\begin{tabular}[#1]{@{}c@{}}#2\end{tabular}}

\usepackage{graphicx}
\usepackage{subcaption}
\usepackage[T1]{fontenc}
\usepackage{makecell}

\graphicspath{{Figures/}}
\usepackage{wrapfig}

\usepackage{amsmath,amssymb,amsfonts} 
\usepackage{fixltx2e}
\usepackage{multirow}
\usepackage{booktabs}

\usepackage{etoolbox}

\usepackage{xcolor}
\usepackage{soul}

\definecolor{mycolor}{rgb}{0.75, 0.75, 0.75}
\definecolor{lavendergray}{rgb}{0.77, 0.76, 0.82}
\definecolor{lightgray}{rgb}{0.83, 0.83, 0.83}
\newcommand{\highlightbox}[1]{\colorbox{mycolor}{$#1$}}

\usepackage{colortbl}

\definecolor{Gray}{gray}{0.38}
\definecolor{LightCyan}{rgb}{0.48,0.78,1}
\definecolor{LightColor1}{rgb}{0.68,0.88,1.2}

\definecolor{LightGray}{gray}{.8}
\newcolumntype{C}[1]{>{\centering\arraybackslash}p{#1}}

\newcolumntype{y}{>{\columncolor{yellow}}c}
\newcolumntype{a}{>{\columncolor{lightgray}}c}
\newcolumntype{g}{>{\columncolor{pink}}c}
\newcolumntype{o}{>{\columncolor{LightColor1}}c}
\newcolumntype{u}{>{\columncolor{LightCyan}}c}

\usepackage{threeparttable}

\begin{document}

\title{
ProSelfLC: 
Progressive Self Label Correction for 
\\Training Robust Deep Neural Networks
\vspace{-0.35cm}
}

\author{
	\fontsize{9.5pt}{1pt}\selectfont
	\vspace{-0.10cm}
	Xinshao Wang\textsuperscript{1, 2},
	~Yang Hua\textsuperscript{3}, 
	~Elyor Kodirov\textsuperscript{1}, 
	~David A. Clifton\textsuperscript{2, }\thanks{Prof. David A. Clifton was supported by the National Institute for Health Research (NIHR) Oxford Biomedical Research Centre (BRC).
	}, 
	~~Neil M. Robertson\textsuperscript{1, 3} \\
	
	\fontsize{9.5pt}{1pt}\selectfont
	\vspace{-0.10cm}
	\textsuperscript{1}Zenith Ai, UK\\
	
	\fontsize{9.5pt}{1pt}\selectfont
	\vspace{-0.10cm}
	\textsuperscript{2}Institute of Biomedical Engineering, University of Oxford, UK\\
	
	\fontsize{9.5pt}{1pt}\selectfont
	\vspace{-0.10cm}
	\textsuperscript{3}Institute of Electronics, Communications and Information Technology, Queen's University Belfast, UK \\
	
	\fontsize{9.5pt}{1pt}\selectfont
	\vspace{-0.10cm}
	\{xinshao, elyor\}@zenithai.co.uk, \{y.hua, n.robertson\}@qub.ac.uk, \{xinshao.wang, david.clifton\}@eng.ox.ac.uk 
	
	\vspace{-0.35cm}
}

\maketitle

\begin{abstract}
	


	
	To train robust deep neural networks (DNNs), we systematically study several target modification approaches, which include output regularisation, self and non-self label correction (LC).  
	%
	Two key issues are discovered:  
	(1) Self LC is the most appealing as it exploits its own knowledge and requires no extra models.
	However, how to automatically decide the trust degree of a learner as training goes is not well answered in the literature?    
	(2) {Some methods penalise while the others reward low-entropy predictions, prompting us to ask which one is better?}
	
	%
	To resolve the first issue, taking two well-accepted propositions--deep neural networks learn meaningful patterns before fitting noise \cite{arpit2017closer} and minimum entropy regularisation principle \cite{grandvalet2006entropy}--we propose a novel end-to-end method named ProSelfLC, which is designed according to learning time and entropy. 
	Specifically, given a data point, we progressively increase trust in its predicted label distribution versus its annotated one if a model has been trained for enough time and the prediction is of low entropy (high confidence).
	%
	%
	For the second issue, according to ProSelfLC, we empirically prove that it is better to redefine a meaningful low-entropy status and optimise the learner toward it. This serves as a defence of entropy minimisation.    
	%
	%
	
	We demonstrate the effectiveness of ProSelfLC through extensive experiments in both clean and noisy settings.   
	The source code is available at 
	\url{https://github.com/XinshaoAmosWang/ProSelfLC-CVPR2021}.

	\vspace{-0.1cm}	
\end{abstract}

\vspace{-0.1cm}
\section{Introduction}
\label{introduction}
\vspace{-0.1cm}

\begin{figure*}[!t]
	\centering
	\vspace{-0.10cm}
	\begin{subfigure}[h!]{\textwidth}
		\centering
		\includegraphics[clip, trim=0cm 2.8cm 0.2cm 1.42cm, width=0.85\textwidth]{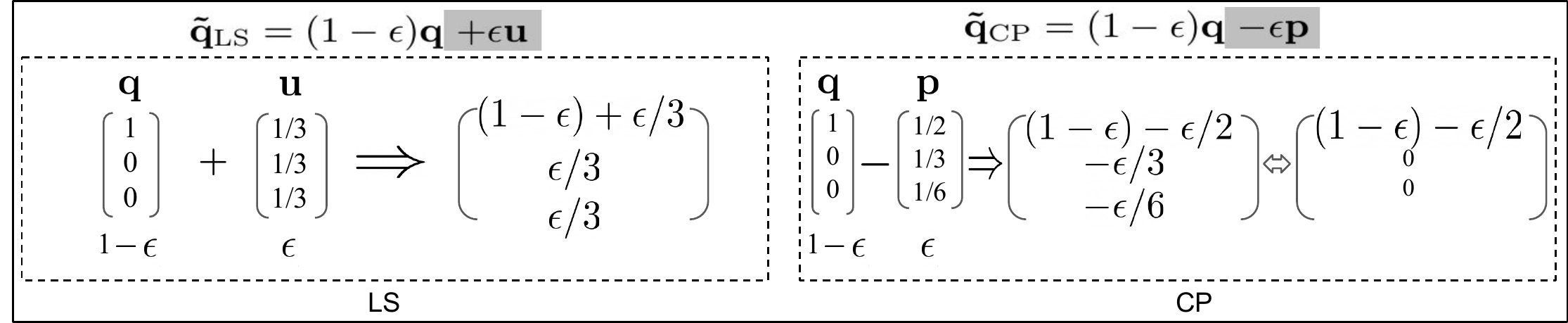}
		\vspace{-0.06cm}
		\caption{OR includes LS \cite{szegedy2016rethinking}
			and CP  
			\cite{pereyra2017regularizing}. 
			LS softens a target by adding a uniform label distribution. 
			CP changes the probability 1 to a smaller value $1-\epsilon$ in the one-hot target. 
			The double-ended arrow means factual equivalence, because an output is definitely non-negative after a softmax layer. 
		}
		\label{fig:LS_CP}
	\end{subfigure}
	
	\begin{subfigure}[h!]{\textwidth}
		\centering
		\includegraphics[clip, trim=-0.2cm 3.2cm 0.3cm 0.8cm, width=0.85\textwidth]{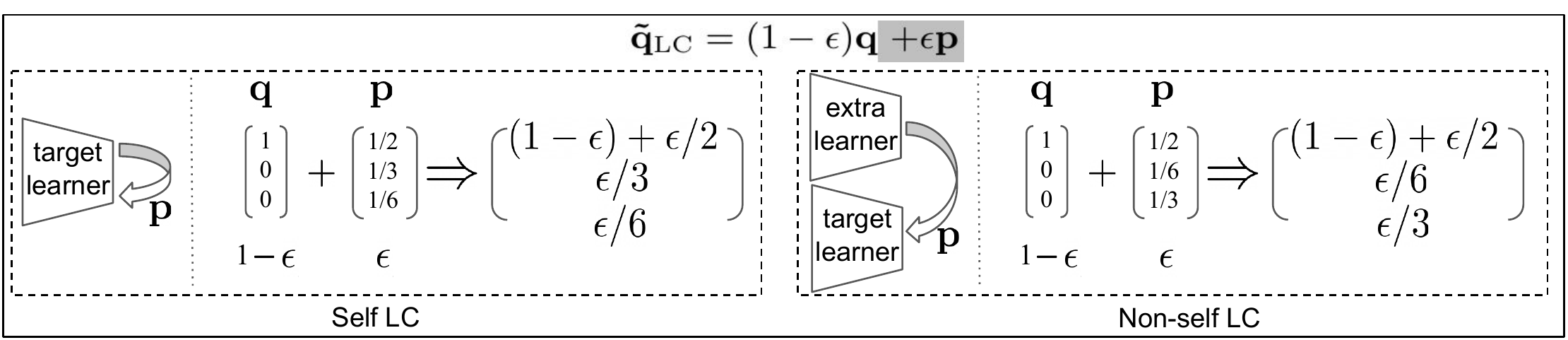}
		\vspace{-0.06cm}
		\caption{LC contains Self LC \cite{lee2013pseudo,reed2015training,tanaka2018joint,yuan2020revisiting}  and Non-self LC \cite{hinton2015distilling}.
			%
			%
			The parameter $\epsilon$ defines how much a predicted label distribution is trusted. 
		}
		\label{fig:LC}
	\end{subfigure}

	\vspace{-0.15cm}
	\caption{Target modification includes OR (LS and CP), and LC (Self LC and Non-self LC).  
		Assume there are three training classes. 
		$\mathbf{q}$ is the one-hot target. 
		$\mathbf{u}$ is a uniform label distribution. 
		$\mathbf{p}$ denotes a predicted label distribution. 
		The	target combination parameter 
		is $\epsilon \in [0,1]$. 
	}
	\label{fig:illustration_LS_CP_LC}
	\vspace{-0.38cm}
\end{figure*}

There exist many target (label) modification approaches. 
They can be roughly divided into two groups: (1) 
%
Output regularisation (OR), which is proposed to penalise overconfident predictions for regularising deep neural networks. It includes label smoothing (LS) \cite{szegedy2016rethinking,muller2019does} and confidence penalty (CP) \cite{pereyra2017regularizing};
(2) Label correction (LC). On the one hand, LC regularises neural networks by adding the similarity structure information over training classes into one-hot label distributions so that the learning targets become \textit{structured and soft}.   
On the other hand, it can \textit{correct the semantic classes} of noisy label distributions. 
%
%
%
%
%
%
%
LC can be further divided into two subgroups: Non-self LC and Self LC. The former requires extra learners, while the latter relies on the model itself. 
A typical approach of Non-self LC is knowledge distillation (KD), which exploits the predictions of other model(s), usually termed teacher(s) \cite{hinton2015distilling}. Self LC methods include Pseudo-Label \cite{lee2013pseudo}, bootstrapping (Boot-soft and Boot-hard) \cite{reed2015training}, Joint Optimisation (Joint-soft and Joint-hard) \cite{tanaka2018joint}, and Tf-KD$_{self}$ \cite{yuan2020revisiting}. 
%
%
According to an overview in Figure~\ref{fig:illustration_LS_CP_LC} (detailed derivation is in Section~\ref{section:preliminary} and Table~\ref{table:summary_CE_LS_CP_LC}), \textit{in label modification,  
	the output target of a data point is defined by combining a one-hot label distribution and its corresponding prediction or a predefined label distribution}. 

%
%
%
%

Firstly, we present the drawbacks of existing approaches: 
(1) OR methods naively penalise confident outputs without leveraging easily accessible knowledge from other learners or itself (Figure \ref{fig:LS_CP}); 
(2) Non-self LC relies on accurate auxiliary models to generate predictions (Figure \ref{fig:LC}). 
(3) Self LC is the most appealing because it exploits its own knowledge and requires no extra learners. 
However, 
there is a core question that is not well answered:
\begin{center}
	\vspace{-0.16cm}
	\textit{In Self LC, how much should we trust a learner to leverage its knowledge?}
	\vspace{-0.15cm}
\end{center}
\noindent
As shown in Figure \ref{fig:LC}, in Self LC, for a data point, we have two labels: a predefined one-hot $\mathbf{q}$ and a predicted structured $\mathbf{p}$. Its learning target is $(1-\epsilon)\mathbf{q}+\epsilon \mathbf{p}$, i.e., a trade-off between $\mathbf{q}$ and $\mathbf{p}$, where $\epsilon$ defines the trust score of a learner. 
%
%
In existing methods, $\epsilon$ is fixed without considering that a model's knowledge grows as the training progresses.
For example,    
in bootstrapping, 
$\epsilon$
is fixed throughout the training process.  
%
Joint Optimisation 
stage-wisely trains a model. It fully trusts predicted labels and uses them to replace old ones when a stage ends, i.e.,  $\epsilon=1$. 
Tf-KD$_{self}$ trains a model by two stages: $\epsilon=0$ in the first one while $\epsilon$ is tuned for the second stage. 
Note that $\mathbf{p}$ is generated by a preceding-stage model in stage-wise training, which requires significant human intervention and is time-consuming in practice. 







To improve Self LC, 
we propose a novel method named Progressive Self Label Correction (ProSelfLC), which is end-to-end trainable and needs negligible extra cost. 
Most importantly, ProSelfLC modifies the target progressively and adaptively as training goes. 
\textit{Two design principles of ProSelfLC are}:  (1) When a model learns from scratch, human annotations are more reliable than its own predictions in the early phase, during which the model is learning simple meaningful patterns before fitting noise, even when severe label noise exists in human annotations \cite{arpit2017closer}. 
(2) As a learner attains confident knowledge as time progresses, we leverage it to revise annotated labels.   
This is surrounded by minimum entropy regularisation, which is widely evaluated in unsupervised and semi-supervised scenarios \cite{grandvalet2005semi,grandvalet2006entropy}. 

%


Secondly, note that OR methods penalise low entropy while LC rewards it, intuitively leading to a second vital question:  
\begin{center}
	\vspace{-0.25cm}
	\textit{Should we penalise a low-entropy status or reward it?}
	\vspace{-0.15cm}
\end{center}
\noindent
Entropy minimisation is the most widely used principle in machine learning \cite{hartigan1979algorithm,rumelhart1986learning,grandvalet2005semi,grandvalet2006entropy,lecun2015deep}. In standard classification, minimising categorical cross entropy (CCE) optimises a model towards a low-entropy status defined by human annotations, which contain noise in very large-scale machine learning.   
As a result, confidence penalty becomes popular for reducing noisy fitting. 
In contrast, we prove that it is better to reward a meaningful low-entropy status redefined by our ProSelfLC. 
Therefore, our work offers a defence of entropy minimisation against the recent confidence penalty practice \cite{szegedy2016rethinking,muller2019does,pereyra2017regularizing,dubey2018maximum}.   

Finally, we summarise our main contributions: 
\begin{itemize}
	\vspace{-0.2cm}
	\item We provide a theoretical study on popular {target modification methods} through entropy and KL divergence \cite{kullback1951information}.  
	Accordingly, we reveal their drawbacks and propose ProSelfLC as a solution. ProSelfLC can: 
	(1) enhance the similarity structure information over training classes; 
	(2) correct the semantic classes of noisy label distributions. 
	ProSelfLC is the first method to trust self knowledge progressively and adaptively. 
	
	\vspace{-0.2cm}
	\item Our extensive experiments: (1) defend the entropy minimisation principle; (2) demonstrate the effectiveness of ProSelfLC in both clean and noisy settings.
	
\end{itemize}

\vspace{-0.22cm}
\section{Related Work}
\label{section:related_work}
\vspace{-0.10cm}

\textbf{\textit{Label noise and semi-supervised learning.}}
%
We test target modification approaches in the setting of label noise because it is generic and connected with semi-supervised learning, 
where only a subset of training examples are annotated, leading to \textit{missing labels}. Then the key to semi-supervised training is to reliably fill them. \textit{When these missing labels are incorrectly filled, the challenge of semi-supervised learning changes to noisy labels.}  
For a further comparison, in semi-supervised learning, the annotated set is clean and reliable, because the label noise only exists in the unannotated set. 
While in our experimental setting, we are not given information on whether an example is trusted or not, thus being even more challenging.      
%
We summarise existing approaches for solving label noise: 
(1) Loss correction, in which we are given  or we need to estimate a noise-transition matrix, which defines the distribution of noise labels \cite{li2017learning,goldberger2017training,sukhbaatar2014learning,vahdat2017toward,yao2019safeguarded,han2018masking,patrini2017making,xiao2015learning}. A noise-transition matrix is difficult and complex to estimate in practice; 
(2) Exploiting an auxiliary trusted training set to differentiate examples \cite{veit2017learning,lee2018cleannet,hendrycks2018using}. 
This requires extra annotation cost; 
(3) Co-training strategies, which train two or more learners \cite{malach2017decoupling,jiang2018mentornet,han2018co,yu2019does,wei2020combating,qiao2018deep,li2020dividemix} and exploit their `disagreement' information to differentiate data points;   
(4) Label engineering methods \cite{song2019selfie,lee2013pseudo,reed2015training,tanaka2018joint,yao2019safeguarded,li2020dividemix}, which relate to our focus in this work.   
Their strategy is to annotate unlabelled samples or correct noisy labels. 

\textit{\textbf{LC and knowledge distillation (KD)}} \cite{bucila2006model,hinton2015distilling}.
%
%
Mathematically, we derive that some KD methods also modify labels. 
We use the term label correction instead of KD for two reasons: (1) label correction is more descriptive; (2) the scope of KD is not limited to label modification.  
For example, multiple networks are trained for KD \cite{furlanello2018born}. 
When two models are trained, the consistency between their predictions of a data point is promoted in \cite{ba2014deep,zhang2018deep}, while the distance between their feature maps is reduced in \cite{romero2015fitnets}.   
Regarding self KD, two examples of the same class are constrained to have consistent output distributions \cite{xu2019data,yun2020regularizing}.  
In another self KD \cite{zhang2019your}, the deepest classifier provides knowledge for shallower classifiers. In a recent self KD method \cite{yuan2020revisiting}, Tf-KD$_{self}$ applies two-stage training. In the second stage, a model is trained by exploiting its knowledge learned in the first stage.  
%
%
Our focus is to improve the end-to-end self LC.  
First, self KD methods \cite{xu2019data,yun2020regularizing,zhang2019your} maximise the consis-
tency of intraclass images’ predictions or the consistency of
different classifiers. In our view, they do not modify labels,
thus being less relevant for comparison. 
Second, the two-stage self KD method \cite{yuan2020revisiting} can be an add-on (i.e., an enhancement plugin) other than a competitor. E.g., in real-world
practice, the first stage can be ProSelfLC instead of CCE with early stopping.
%
%
Finally, we acknowledge that exploiting ProSelfLC to improve non-self KD and stage-wise approaches is an important area for future work, e.g., a better teacher model can be trained using ProSelfLC.

%
%

\vspace{-0.10cm}
\section{Mathematical Analysis and Theory}
\label{section:preliminary}
\vspace{-0.10cm}


Let $\mathbf{X}=\{(\mathbf{x}_i, y_i)\}_{i=1}^N$ represent $N$ training examples, where $(\mathbf{x}_i, y_i)$ denotes $i-$th sample with input $\mathbf{x}_i \in \mathbb{R}^D$ and label $y_i \in \{1,2, ..., C\}$. $C$ is the number of classes. A deep neural network $z$ consists of an embedding network $f(\cdot): \mathbb{R}^D \rightarrow \mathbb{R}^K$ and a linear classifier $g(\cdot): \mathbb{R}^K \rightarrow \mathbb{R}^C$, i.e., $\mathbf{z}_i=z(\mathbf{x}_i)=g(f(\mathbf{x}_i)): \mathbb{R}^D \rightarrow \mathbb{R}^C$.
For the brevity of analysis, we take a data point and omit its subscript so that it is denoted by $(\mathbf{x}, y)$.  
The linear classifier is usually the last fully-connected layer. Its output is named logit vector $\mathbf{z} \in \mathbb{R}^C$. 
We produce its classification probabilities $\mathbf{p}$ by normalising the logits using a softmax function: 
\begin{equation}
	\begin{aligned}
		\mathbf{p}(j|\mathbf{x})= 
		{\exp (\mathbf{z}_{j})}
		/{ \sum\nolimits_{m=1}^{C} \exp (\mathbf{z}_{m}) },
	\end{aligned}
	\label{chapter:DM_eq:softmax_normalisation}
\end{equation} 
where $\mathbf{p}(j|\mathbf{x})$ is the probability of $\mathbf{x}$ belonging to class $j$. 
Its corresponding ground-truth is usually denoted by a one-hot representation $\mathbf{q}$: $\mathbf{q}(j|\mathbf{x})=1$ if $j=y$,  $\mathbf{q}(j|\mathbf{x})=0$ otherwise.

\vspace{-0.10cm}
\subsection{Semantic class and similarity structure in a label distribution}
\vspace{-0.10cm}

A probability vector $\mathbf{p} \in \mathbb{R}^C$ can also be interpreted as an instance-to-classes similarity vector, i.e., $\mathbf{p}(j|\mathbf{x})$ measures how much a data point $\mathbf{x}$ is similar with (analogously, likely to be) $j$-th class. Consequently, $\mathbf{p}$ should not be exactly one-hot, and is proposed to be corrected at training, so that it can define a more informative and structured learning target. For better clarity, we first present two definitions: 

\textbf{Definition 1} (\textit{Semantic Class}). Given a target label distribution $\mathbf{\tilde{q}}(\mathbf{x}) \in \mathbb{R}^C$,  the semantic class is defined by $\argmax\nolimits_j {\mathbf{{\tilde{q}}}(j|\mathbf{x})}$, i.e., the class whose probability is the largest.  

\textbf{Definition 2} (\textit{Similarity Structure}). 
In $\mathbf{\tilde{q}}(\mathbf{x})$, $\mathbf{x}$ has $C$ probabilities of being predicted to $C$ classes. The similarity structure of $\mathbf{x}$ versus $C$ classes is defined by these probabilities and their differences. 

\vspace{-0.10cm}
\subsection{Revisit of CCE, LS, CP and LC}
\vspace{-0.10cm}

\textbf{Standard CCE}.
For any input $(\mathbf{x}, y)$, the minimisation objective of standard CCE is: 
\vspace{-0.12cm}
\begin{equation}
	\label{eq:cross_entropy}
	\vspace{-0.12cm}
	\begin{aligned}
		L_{\mathrm{CCE}} ( \mathbf{q}, \mathbf{p} )  =
		\mathrm{H}(\mathbf{q}, \mathbf{p}) 
		&= \mathrm{E}_\mathbf{q} ( -\log~\mathbf{p} ), 
	\end{aligned}
\end{equation}
where $\mathrm{H(\cdot, \cdot)}$ represents the cross entropy. $\mathrm{E}_\mathbf{q} ( \mathbf{-\log~\mathbf{p}} ) $ denotes the expectation of negative log-likelihood, and $\mathbf{q}$ serves as the probability mass function.


\textbf{Label smoothing}.
In LS \cite{szegedy2016rethinking,hinton2015distilling}, 
we soften one-hot targets by adding a uniform distribution:
$\mathbf{\tilde{q}_{\mathrm{LS}}} = (1-\epsilon)\mathbf{q}+\epsilon \mathbf{u}$, $\mathbf{u} \in \mathbb{R}^C, \text{ and } \forall j,  \mathbf{u}_j = \frac{1}{C}  $.
Consequently:
\vspace{-0.12cm}
\begin{equation}
	\label{eq:label_smoothing}
	\vspace{-0.12cm}
	\begin{aligned}
		L_\mathrm{{CCE+LS}}(\mathbf{{q}}, \mathbf{p}; \epsilon) 
		&=
		\mathrm{H}(\mathbf{\tilde{q}_{\mathrm{LS}}}, \mathbf{p}) 
		= \mathrm{E}_\mathbf{\tilde{q}_{\mathrm{LS}}} ( -\log~\mathbf{p} ) 
		\\&
		= (1-\epsilon) \mathrm{H}(\mathbf{q}, \mathbf{p})
		{+
			\epsilon \mathrm{H}(\mathbf{u}, \mathbf{p}).}
	\end{aligned}
\end{equation}

\textbf{Confidence penalty}.
CP \cite{pereyra2017regularizing} penalises highly confident predictions:  
\vspace{-0.12cm}
\begin{equation}
	\label{eq:confidence_penalty}
	\vspace{-0.12cm}
	\begin{aligned}
		L_\mathrm{{CCE+CP}}(\mathbf{{q}}, \mathbf{p}; \epsilon) 
		&
		= (1-\epsilon) \mathrm{H}(\mathbf{q}, \mathbf{p})
		{-
			\epsilon \mathrm{H}(\mathbf{p}, \mathbf{p})}. 
	\end{aligned}
\end{equation}
%
%
%
%
%
%
%
%

\textbf{Label correction}.
As illustrated in Figure~\ref{fig:illustration_LS_CP_LC}, LC is a family of algorithms, where a one-hot label distribution is modified to a convex combination of itself and a predicted distribution: 
\vspace{-0.12cm}
\begin{equation}
	\label{eq:label_correction}
	\vspace{-0.12cm}
	\begin{aligned}
		 \mathbf{\tilde{q}_{\mathrm{LC}}} = (1-\epsilon)\mathbf{q}+\epsilon \mathbf{p}
		~\Rightarrow~
		&L_\mathrm{{CCE+LC}}(\mathbf{{q}}, \mathbf{p}; \epsilon) 
		= \mathrm{H}(\mathbf{\tilde{q}_{\mathrm{LC}}}, \mathbf{p})  
		\\&= (1-\epsilon) \mathrm{H}(\mathbf{q}, \mathbf{p})
		{+
			\epsilon \mathrm{H}(\mathbf{p}, \mathbf{p})}.
	\end{aligned}
\end{equation}
We remark: (1) $\mathbf{p}$ provides meaningful information about an example's relative probabilities of being different training classes; (2) If $\epsilon$ is large, and $\mathbf{p}$ is confident in predicting a different class, i.e.,  $\argmax\nolimits_j \mathbf{p}(j|\mathbf{x}) \neq \argmax\nolimits_j \mathbf{q}(j|\mathbf{x})$, $\mathbf{\tilde{q}_{\mathrm{LC}}}$ defines a different semantic class from $\mathbf{q}$.

\begin{table*}[!t]
	\vspace{-0.10cm}
	\caption{
		Summary of CCE, LS, CP and LC. 
	}
	\centering
	\setlength{\tabcolsep}{10.0pt} 
	\fontsize{9.5pt}{9.5pt}\selectfont

	\vspace{-0.32cm}
	\begin{tabular}{l c c c c}
		\toprule
		&  \multicolumn{1}{c}{CCE}
		&  \multicolumn{1}{c}{LS}
		
		& \multicolumn{1}{c}{CP}
		
		& \multicolumn{1}{c}{LC}
		\\
		\midrule
		{Learning Target}
		&  $\mathbf{q}$ & $ \mathbf{\tilde{q}_{\mathrm{LS}}}=(1-\epsilon)\mathbf{q} \highlightbox{+\epsilon \mathbf{u}}$ & $ \mathbf{\tilde{q}_{\mathrm{CP}}}=(1-\epsilon)\mathbf{q}\highlightbox{-\epsilon \mathbf{p}}$& $ \mathbf{\tilde{q}_{\mathrm{LC}}}=(1-\epsilon)\mathbf{q}\highlightbox{+\epsilon \mathbf{p}}$\\
		Cross Entropy & $\mathrm{E}_\mathbf{q} ( -\log~\mathbf{p} )$
		& $\mathrm{E}_\mathbf{\tilde{q}_{\mathrm{LS}}} ( -\log~\mathbf{p} )$ 
		& $\mathrm{E}_\mathbf{\tilde{q}_{\mathrm{CP}}} ( -\log~\mathbf{p} )$
		& $\mathrm{E}_\mathbf{\tilde{q}_{\mathrm{LC}}} ( -\log~\mathbf{p} )$
		\\
		\vspace{-0.28cm}
		&&&&\\
		\cdashline{1-5}
		\vspace{-0.28cm}
		&&&&\\
		KL Divergence &  $\mathrm{KL}(\mathbf{q}||\mathbf{p})$ & \specialcell{$(1-\epsilon) \mathrm{KL}(\mathbf{q}||\mathbf{p})
			$\\
			$\highlightbox{+\epsilon \mathrm{KL}(\mathbf{u}||\mathbf{p})}$} & 
		\specialcell{$(1-\epsilon) \mathrm{KL}(\mathbf{q}||\mathbf{p})
			$\\
			$\highlightbox{+\epsilon \mathrm{KL}(\mathbf{p}||\mathbf{u})}$} &
		\specialcell{$(1-\epsilon) \mathrm{KL}(\mathbf{q}||\mathbf{p})
			$\\
			$\highlightbox{-\epsilon \mathrm{KL}(\mathbf{p}||\mathbf{u})}$} 
		\\
		
		\vspace{-0.28cm}
		&&&&\\
		\cdashline{1-5}
		\vspace{-0.28cm}
		&&&&\\
		
		Entropy minimisation & -- & \highlightbox{Penalise} over CCE & \highlightbox{Penalise} over CCE & \highlightbox{Reward} over CCE\\

		Semantic class & Annotated & Annotated & Annotated & Annotated and Learned \\
		
		Similarity structure & No & No & No & Yes\\
		
		\bottomrule
	\end{tabular}
	\label{table:summary_CE_LS_CP_LC}
	\vspace{-0.30cm}
\end{table*}

\vspace{-0.1cm}
\subsection{Theory}
\vspace{-0.10cm}

\textbf{Proposition 1.} \textit{LS, CP and LC modify the learning targets of standard CCE.}\\
\textit{Proof.}  
$L_\mathrm{{CCE+CP}}(\mathbf{{q}}, \mathbf{p}; \epsilon) 
= (1-\epsilon) \mathrm{H}(\mathbf{q}, \mathbf{p})
{-
	\epsilon \mathrm{H}(\mathbf{p}, \mathbf{p})}=\mathrm{E}_\mathbf{(1-\epsilon)\mathbf{q}{-\epsilon \mathbf{p}}} ( \mathbf{-\log~\mathbf{p}} )$. Therefore, 
$ \mathbf{\tilde{q}_{\mathrm{CP}}}=(1-\epsilon)\mathbf{q}{-\epsilon \mathbf{p}}$. 
Additionally,   
$ \mathbf{\tilde{q}_{\mathrm{LS}}}=(1-\epsilon)\mathbf{q} {+\epsilon \mathbf{u}}$,  
$\mathbf{\tilde{q}_{\mathrm{LC}}}=(1-\epsilon)\mathbf{q}{+\epsilon \mathbf{p}}$. \hfill\(\Box\)

\textbf{Proposition 2.} \textit{Some KD methods, which aim to minimise the KL divergence between predictions of a teacher and a student, belong to the family of label correction. }\\
\textit{Proof.} In general, a loss function of such methods can be defined to be $L_\mathrm{KD}(\mathbf{q},\mathbf{p}_t, \mathbf{p}) = (1-\epsilon)\mathrm{H}(\mathbf{q}, \mathbf{p}) + \epsilon \mathrm{KL}(\mathbf{p}_t||\mathbf{p})$ \cite{yuan2020revisiting}.  $\mathrm{KL(\cdot||\cdot)}$ denotes the KL divergence. As $\mathrm{KL}(\mathbf{p}_t||\mathbf{p}) = \mathrm{H}(\mathbf{p}_t, \mathbf{p})-\mathrm{H}(\mathbf{p}_t, \mathbf{p}_t)$, $\mathbf{p}_t$ is from a teacher and fixed when training a student. We can omit $\mathrm{H}(\mathbf{p}_t, \mathbf{p}_t)$:
\vspace{-0.12cm}
\begin{equation}
	\label{eq:KD_KLptp}
	\vspace{-0.12cm}
	\begin{aligned}
	\fontsize{9.5pt}{9.5pt}\selectfont
	L_\mathrm{KD}(\mathbf{q},\mathbf{p}_t, \mathbf{p}) 
	&= (1-\epsilon)\mathrm{H}(\mathbf{q}, \mathbf{p}) + \epsilon \mathrm{H}(\mathbf{p}_t,\mathbf{p})
	\\&= \mathrm{E}_{(1-\epsilon)\mathbf{q}{+\epsilon \mathbf{p}_t}} ( \mathbf{-\log~\mathbf{p}} ) \\&\Rightarrow \mathbf{\tilde{q}_{\mathrm{KD}}} = (1-\epsilon)\mathbf{q}+\epsilon \mathbf{p}_t .
	\end{aligned}
\end{equation}
Consistent with LC in Eq~(\ref{eq:label_correction}), $L_\mathrm{KD}(\mathbf{q},\mathbf{p}_t, \mathbf{p})$ revises a label using $\mathbf{p}_t$. \hfill\(\Box\)

\textbf{Proposition 3.} \textit{Compared with CCE, LS and CP penalise entropy minimisation while LC reward it.}

\textbf{Proposition 4.} \textit{In CCE, LS and CP, a data point $\mathbf{x}$ has the same semantic class. In addition, $\mathbf{x}$ has an identical probability of belonging to other classes except for its semantic class.
}
%
%

The proof of propositions 3 and 4 is presented in the Appendix \ref{appendix_sec:proof_of_propositions}. 
Only LC exploits informative information and has the ability to correct labels, while LS and CP only relax the hard targets.
We summarise CCE, LS, CP and LC in Table~\ref{table:summary_CE_LS_CP_LC}. Constant terms are ignored for concision.



\begin{table*}[!t]
	\vspace{-0.10cm}
	\caption{
		The values of  $g(t)$, $l(\mathbf{p})$ and $\epsilon_{\mathrm{ProSelfLC}} = g(t) \times l(\mathbf{p})$ under different cases. 
		We use concrete values for concise interpretation.   
		We bold the special case when the semantic class is changed.
		%
		%
	}
	\centering
	\setlength{\tabcolsep}{10.0pt} 
	\fontsize{9.5pt}{9.5pt}\selectfont

	\vspace{-0.28cm}
	\begin{tabular}{l c c c c}
		\toprule
		&  & \multicolumn{3}{c}{$l(\mathbf{p})$: 
			Consistency is defined by whether $\mathbf{p}$ and $\mathbf{q}$ share the semantic class or not. 	
		} 
		\\
		
		\cmidrule(l{2pt}r{2pt}){3-5}
		
		~~~~~~~~~~~~ &  &  ~~~0.1(non-confident) & ~~~~0.9(confidently consistent)  & ~~~0.9(confidently inconsistent) \\
		
		\midrule
		{Earlier phase}  & $g(t)=0.1$ & 0.01 & ~~~~0.09 & {0.09}\\
		{Later phase} & $g(t)=0.9$ & 0.09 & ~~~~0.81 & \textbf{0.81}
		\\
		\bottomrule
	\end{tabular}
	\label{table:case_analysis_ProSelfLC}
	\vspace{-0.41cm}
\end{table*}

\begin{figure*}[!t]
	\vspace{-0.42cm}
	\centering
	\begin{subfigure}[!h]{0.33\textwidth}
		\centering
		\captionsetup{width=\textwidth}
		\includegraphics[clip, trim=3.05cm 8.9cm 4.2cm 7.9cm, width=0.99\textwidth]{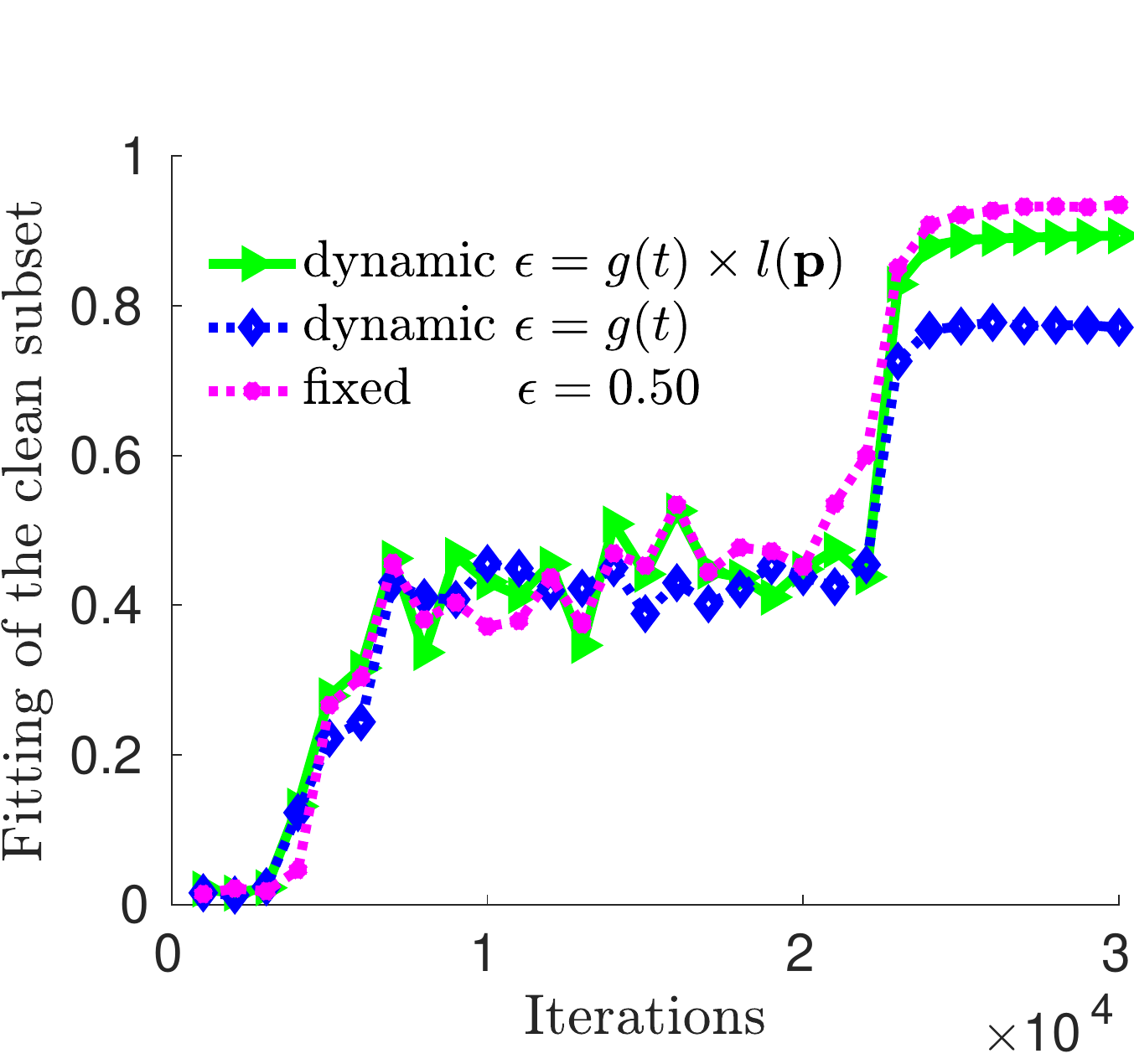}
		\caption{Correct fitting.}
		\label{fig:AblationNoEntropyCIFAR100Asy0_4_cleanFittting}
	\end{subfigure}
	%
	%
	\begin{subfigure}[!h]{0.329\textwidth}
		\centering
		\captionsetup{width=\textwidth}
		\includegraphics[clip, trim=3.05cm 8.9cm 4.2cm 7.9cm, width=0.99\textwidth]{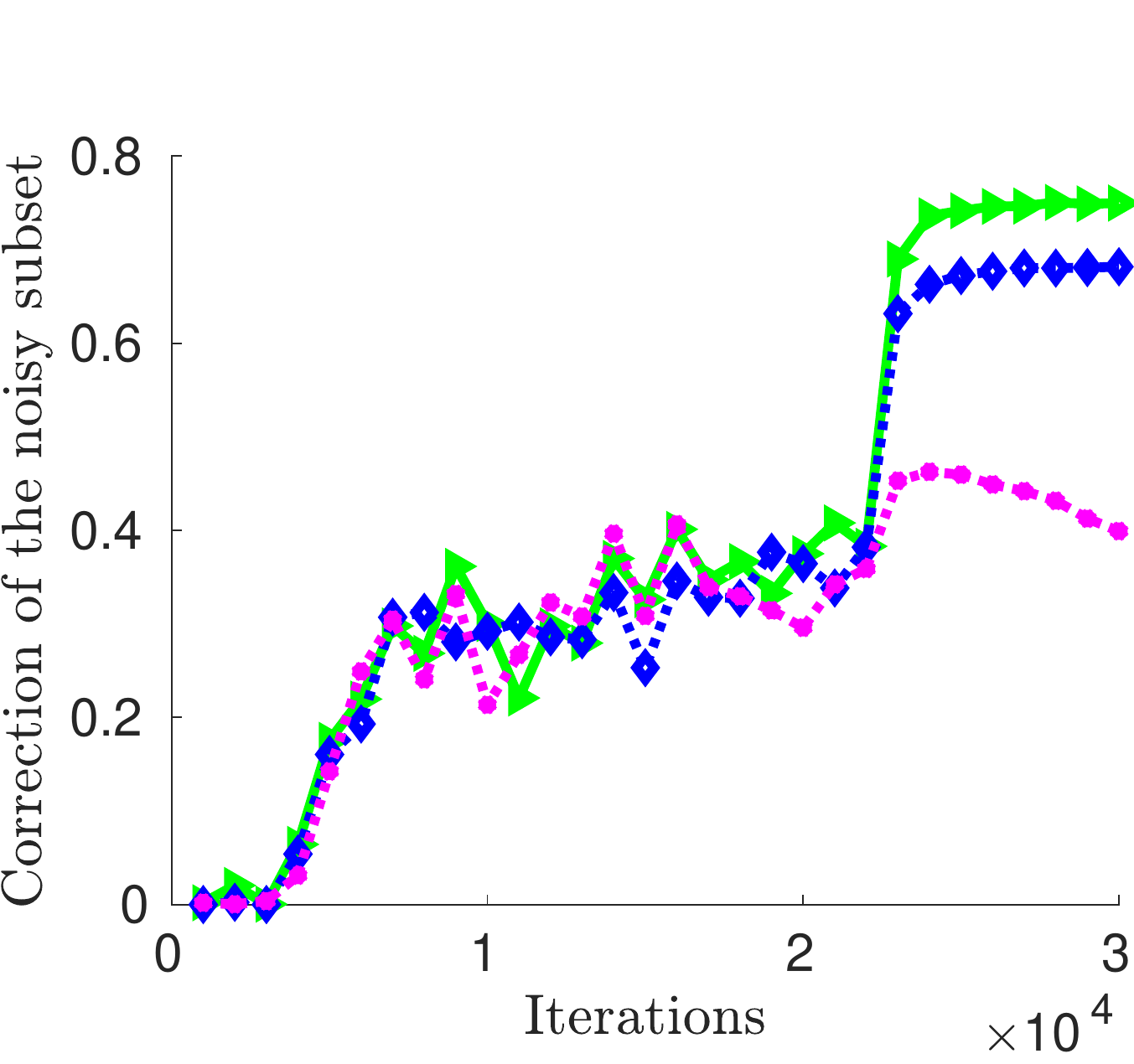}
		\caption{Semantic class correction}
		\label{fig:AblationNoEntropyCIFAR100Asy0_4_semanticCorrection}
	\end{subfigure}
	%
	%
	%
	%
	%
	\begin{subfigure}[!h]{0.329\textwidth}
		\centering
		\captionsetup{width=\textwidth}
		\includegraphics[clip, trim=3.05cm 8.9cm 4.2cm 7.9cm, width=0.99\textwidth]{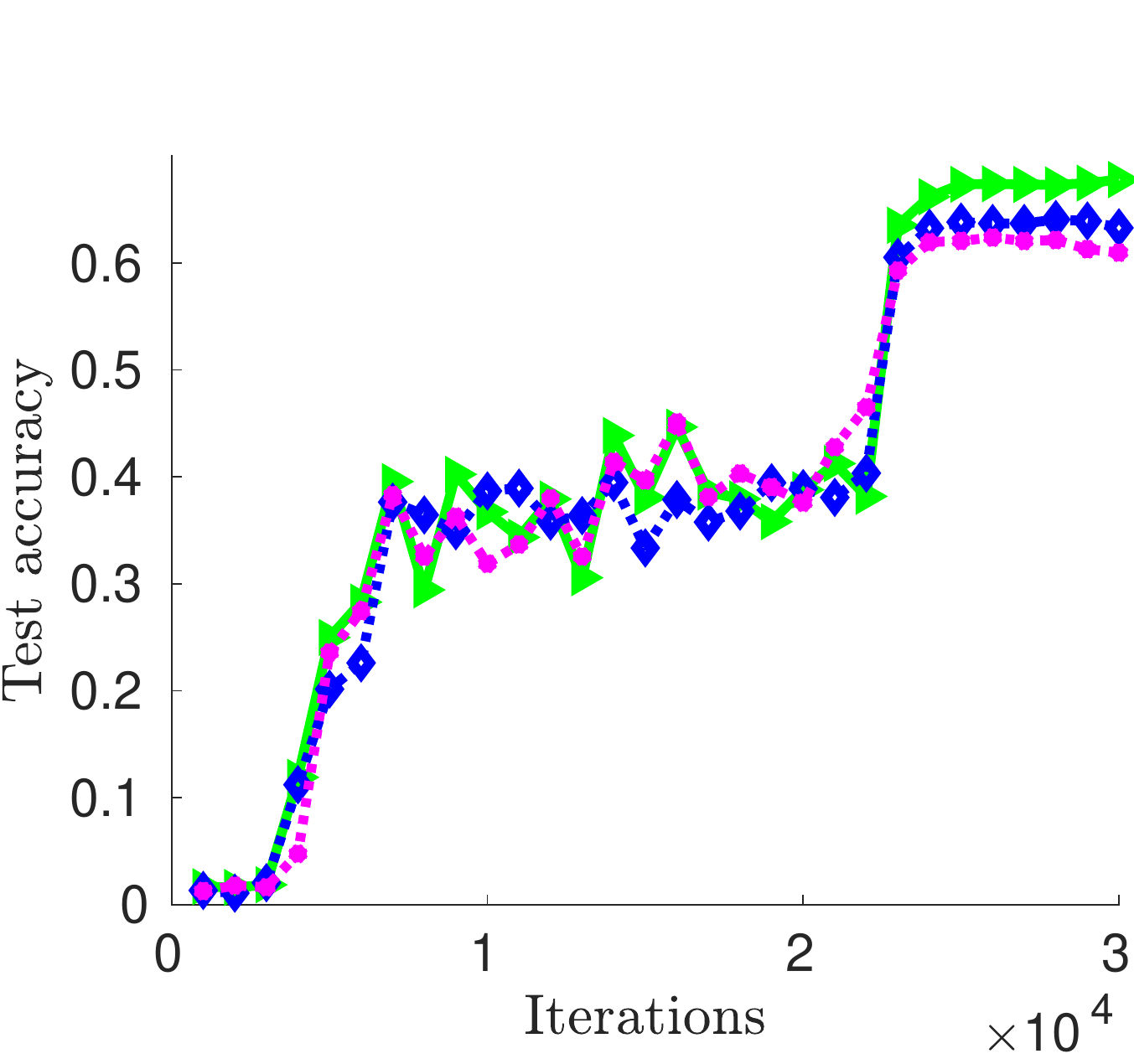}
		\caption{Generalisation.}
		\label{fig:AblationNoEntropyCIFAR100Asy0_4_accuracy30000}
	\end{subfigure}

	\vspace{-0.28cm}
	\caption{ 
		Study of setting $\epsilon$ using three schemes: global trust and local trust, merely global trust, and fixed $\epsilon$. 
		Experiments are done on CIFAR-100 with $40\%$ asymmetric label noise. 
		Vertical axes denote evaluation metrics.    
		Mean results are displayed.   
	}
	\label{fig:AblationNoEntropy}
	\vspace{-0.30cm}
\end{figure*}

\vspace{-0.1cm}
\section{ProSelfLC: Progressive and Adaptive Label Correction 
}
\label{section:proselflc}
\vspace{-0.10cm}

In standard CCE, a semantic class is considered while the similarity structure is ignored. It is mainly due to the difficulty of annotating the similarity structure for every data point, especially when $C$ is large \cite{xu2020variational}. 
Fortunately, recent progress demonstrates that there are some effective approaches to define the similarity structure of data points without annotation: 
(1) In KD, an auxiliary teacher model can provide a student model the similarity structure information \cite{hinton2015distilling,muller2019does}; 
(2) In Self LC, e.g., Boot-soft, a model helps itself by exploiting the knowledge it has learned so far. 
We focus on studying the end-to-end Self LC. 



In Self LC, $\epsilon$ indicates how much a predicted label distribution is trusted.  
In ProSelfLC, we propose to set it automatically according to learning time $t$ and prediction entropy $\mathrm{H}(\mathbf{p})$, i.e., ProSelfLC trusts self knowledge according to training time and confidence. 
For any $\mathbf{x}$, we summarise: 
\vspace{-0.12cm}
\begin{equation}
	\label{eq:ProSelfLC}
	\fontsize{9.5pt}{9.5pt}\selectfont
	\vspace{-0.12cm}
	\begin{aligned}
	\begin{cases}
		\begin{aligned}
		\text{Loss: }
		{L}_( \mathbf{\tilde{q}_{\mathrm{ProSelfLC}}}, \mathbf{p};\epsilon_{\mathrm{ProSelfLC}} )  
		&=\mathrm{H}(\mathbf{\tilde{q}_{\mathrm{ProSelfLC}}}, \mathbf{p})
		\\&=\mathrm{E}_\mathbf{\tilde{q}_{\mathrm{ProSelfLC}}} ( -\log~\mathbf{p} ).
		\end{aligned}
		
		\\~
		
		\\
		\begin{aligned}
		\text{Label: } \mathbf{\tilde{q}_{\mathrm{ProSelfLC}}}
		=(1- \epsilon_{\mathrm{ProSelfLC}})  \mathbf{q}+\epsilon_{\mathrm{ProSelfLC}} \mathbf{p}. 
		\end{aligned}
		
		\\~
		\\
		\begin{aligned}
		\epsilon_{\mathrm{ProSelfLC}} = 
		g(t)
		\times 
		l(\mathbf{p})
		\begin{cases} 
			g(t) = h({t}/{\Gamma}-0.5, \emB) \in (0,1),    \\
			\\~
			l(\mathbf{p}) = 1 - 
			{\mathrm{H}(\mathbf{p})}/{\mathrm{H}(\mathbf{u})}  \in (0,1).
		\end{cases}
		\end{aligned}
	\end{cases}
	\end{aligned}
\end{equation}
$t$ and $\Gamma$ are the iteration counter and the number of total iterations, respectively. 
$h(\eta, \emB)= {1}/({1+\exp(-\eta \times \emB)})$. Here, $\eta = {t}/{\Gamma}-0.5$.  
$\emB, \Gamma$ are task-dependent and searched on a validation set. 


\vspace{-0.1cm}
\subsection{Self trust scores}
\vspace{-0.10cm}

{\textbf{Global trust score} $g(t)$} denotes how much we trust a learner. It is independent of data points, thus being global. $g(t)$ grows as $t$ rises. 
$\emB$ adjusts the exponentiation's base and growth speed of $g(t)$.

\textbf{Local trust score} $l(\mathbf{p})$ indicates how much we trust an output distribution $\mathbf{p}$, which is data-dependent.    
$l(\mathbf{p})$ rises as ${\mathrm{H}(\mathbf{p})}$ becomes lower, rewarding a confident distribution.       


\vspace{-0.10cm} 
\subsection{Design reasons}
\vspace{-0.1cm}

Regarding $g(t)$, 
in the earlier learning phase, i.e., $t < \Gamma/2$, $g(t) < 0.5 \Rightarrow \epsilon_{\mathrm{ProSelfLC}} < 0.5, \forall \mathbf{p}$, so that the human annotations dominate and ProSelfLC only modifies the similarity structure. 
%
%
When a learner has not seen the training data for enough time at the earlier stage, its knowledge is less reliable and a wrong confident prediction may occur.
Our design assuages the bad impact of such unexpected cases.
When it comes to the later training phase, i.e., $t > \Gamma/2$, we have $g(t) > 0.5$
as it has been trained for more than half of entire iterations. 

Regarding $l(\mathbf{p})$, it affects the later learning phase. 
If $\mathbf{p}$ is less confident, $l(\mathbf{p})$ will be smaller, then $\epsilon_{\mathrm{ProSelfLC}}$ will be smaller, hence {we trust $\mathbf{p}$ less when it is of higher uncertainty}. 
If $\mathbf{p}$ is highly confident, we trust its confident knowledge. 
%
\textit{Ablation study of our design is in Figure~\ref{fig:AblationNoEntropy}, where three variants of $\epsilon$ are presented}. In our experiments, note that when $\epsilon$ is fixed, we try three values (0.125, 0.25, 0.50) and display the best instantiation, i.e., $\epsilon=0.50$.
%

\vspace{-0.10cm}
\subsection{Case analysis}
\vspace{-0.1cm}

Due to the potential memorisation in the earlier phase (which is more rare), we may get unexpected  confident wrong predictions for noisy labels, but their trust scores are small as $g(t)$ is small.  
We conduct the case analysis of ProSelfLC in Table~\ref{table:case_analysis_ProSelfLC} and summarise its core tactics as follows:

(1) {{Correct the similarity structure for every data point in all cases}}, thanks to exploiting the self knowledge of a learner, i.e., $\mathbf{p}$. 
%

\begin{figure*}[!t]
	\vspace{-0.48cm}
	\centering
	\begin{subfigure}[!h]{0.33\textwidth}
		\centering
		\captionsetup{width=\textwidth}
		\includegraphics[clip, trim=3.05cm 8.9cm 4.2cm 7.9cm, width=0.99\textwidth]{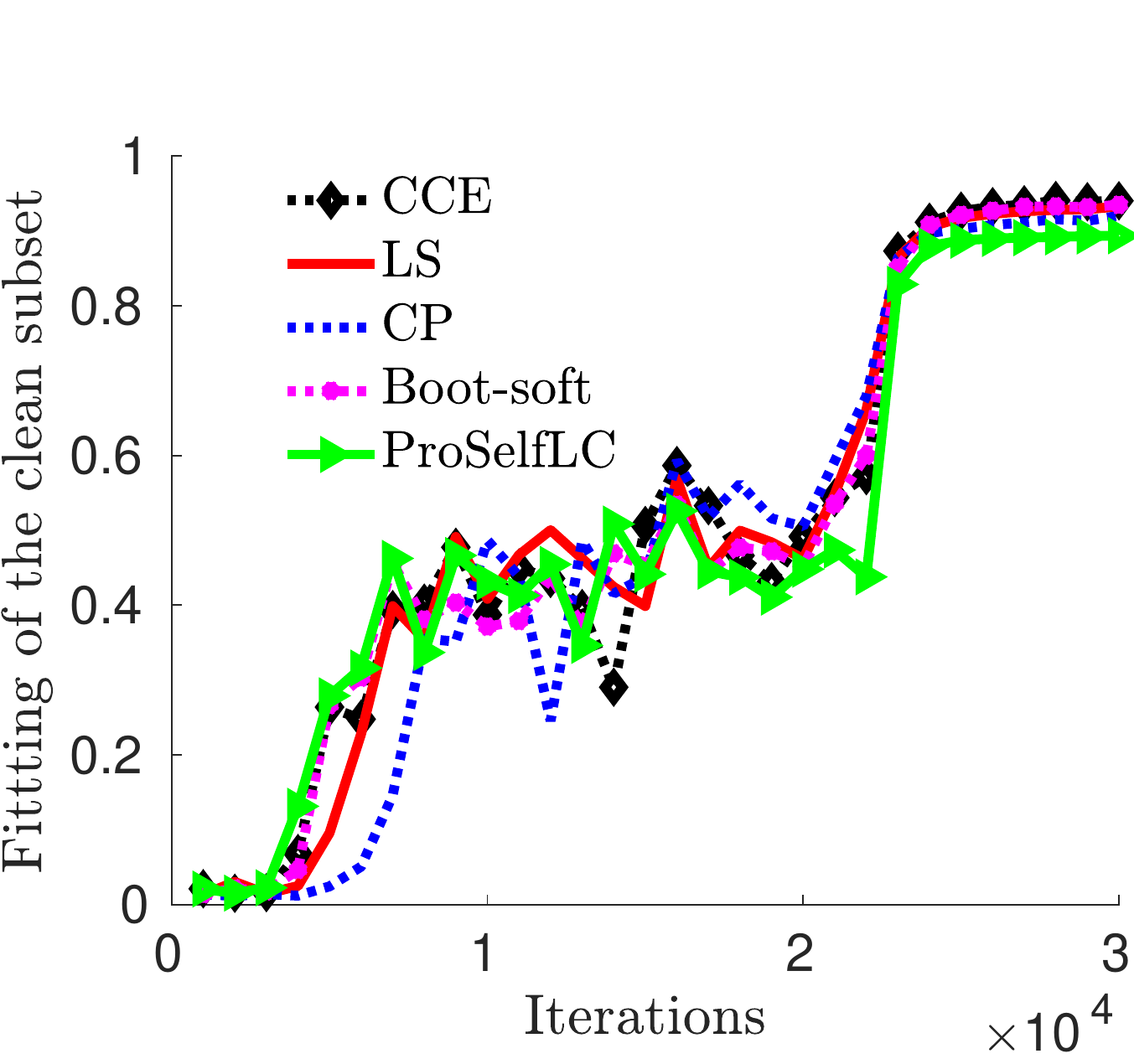}
		\caption{Correct fitting.}
		\label{fig:CIFAR100Asy0_4_cleanFittting}
	\end{subfigure}
	\begin{subfigure}[!h]{0.329\textwidth}
		\centering
		\captionsetup{width=\textwidth}
		\includegraphics[clip, trim=3.05cm 8.9cm 4.2cm 7.9cm, width=0.99\textwidth]{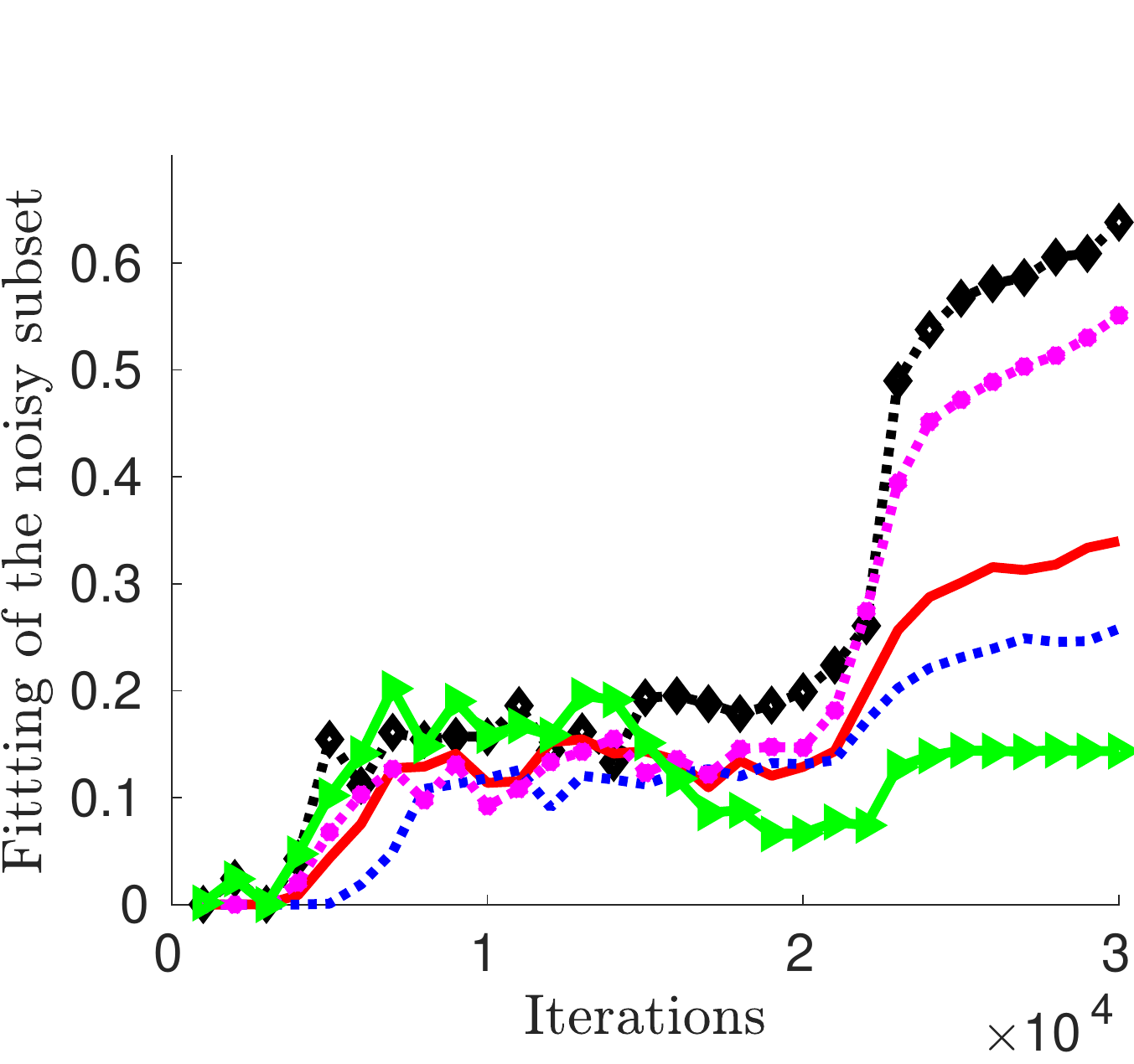}
		\caption{Wrong fitting.}
		\label{fig:CIFAR100Asy0_4_corruptedFittting}
	\end{subfigure}
	\begin{subfigure}[!h]{0.329\textwidth}
		\centering
		\captionsetup{width=\textwidth}
		\includegraphics[clip, trim=3.05cm 8.9cm 4.2cm 7.9cm, width=0.99\textwidth]{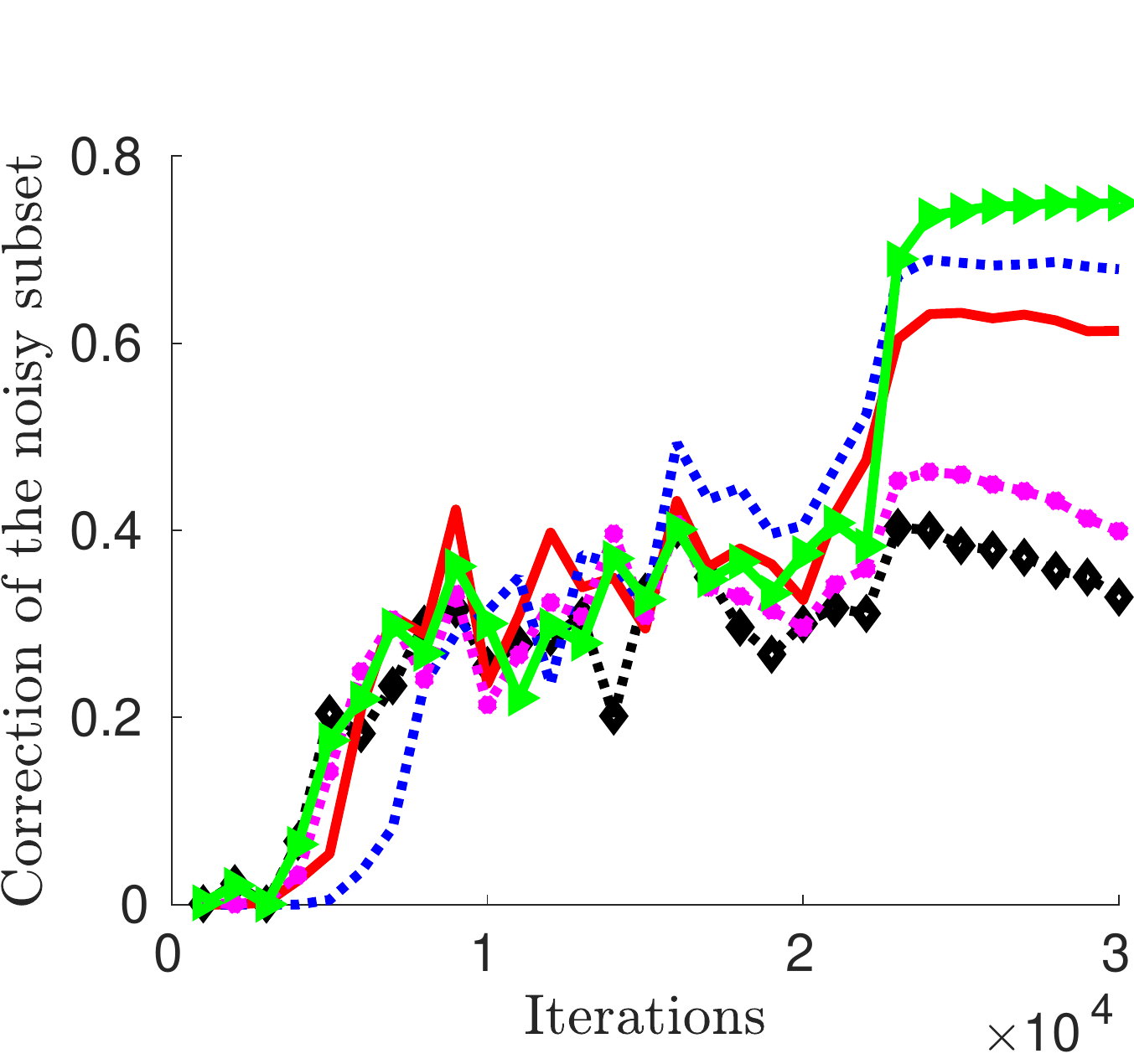}
		\caption{Semantic class correction}
		\label{fig:CIFAR100Asy0_4_semanticCorrection}
	\end{subfigure}
	
	\vspace{-0.39cm}
	\begin{subfigure}[!h]{0.329\textwidth}
		\centering
		\captionsetup{width=\textwidth}
		\includegraphics[clip, trim=3.05cm 8.9cm 4.2cm 7.9cm, width=0.99\textwidth]{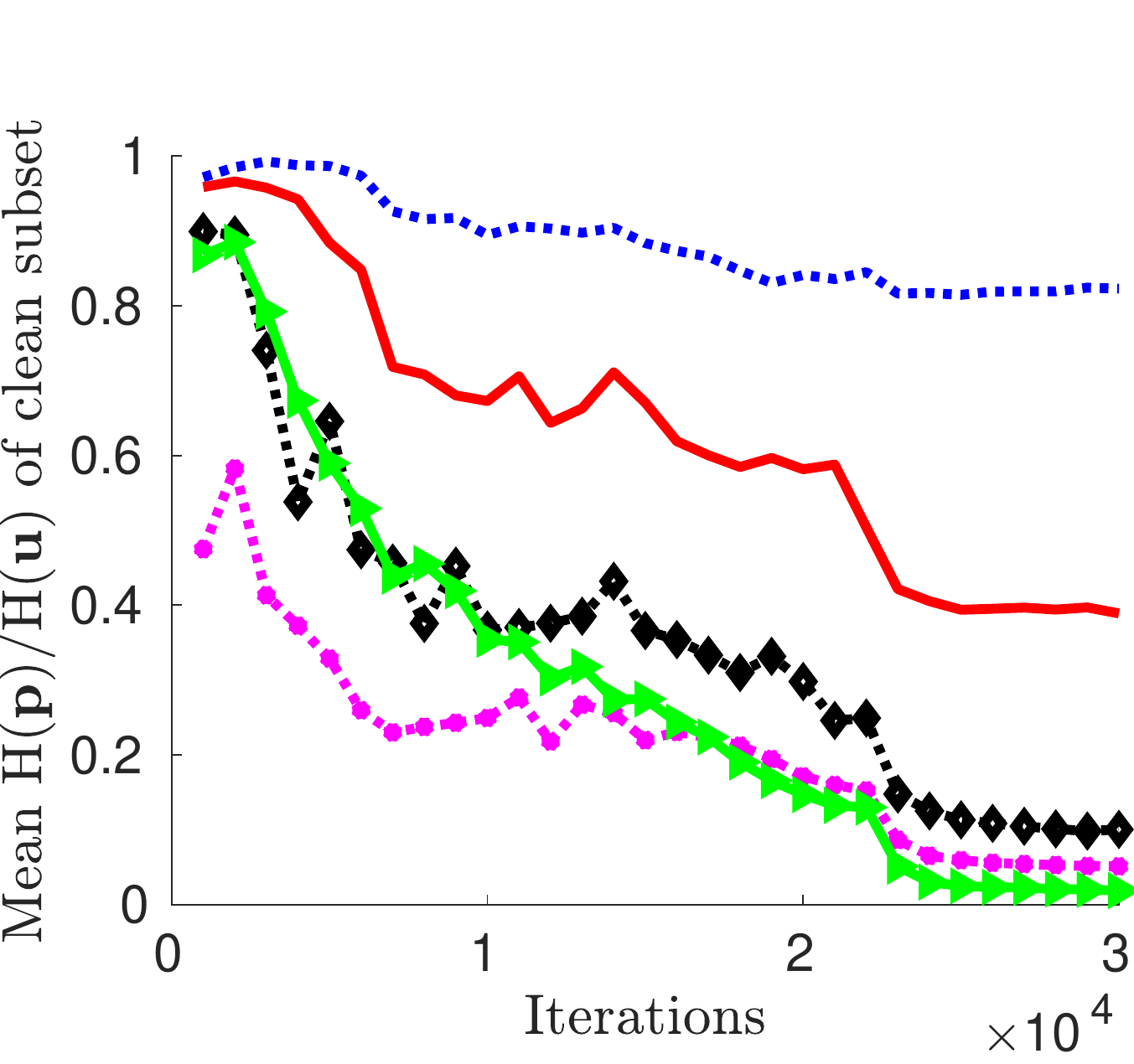}
		\caption{Entropy of clean subset.}
		\label{fig:CIFAR100Asy0_4_mean_entropy_CleanSubset}
	\end{subfigure}
	\begin{subfigure}[!h]{0.329\textwidth}
		\centering
		\captionsetup{width=\textwidth}
		\includegraphics[clip, trim=3.05cm 8.9cm 4.2cm 7.9cm, width=0.99\textwidth]{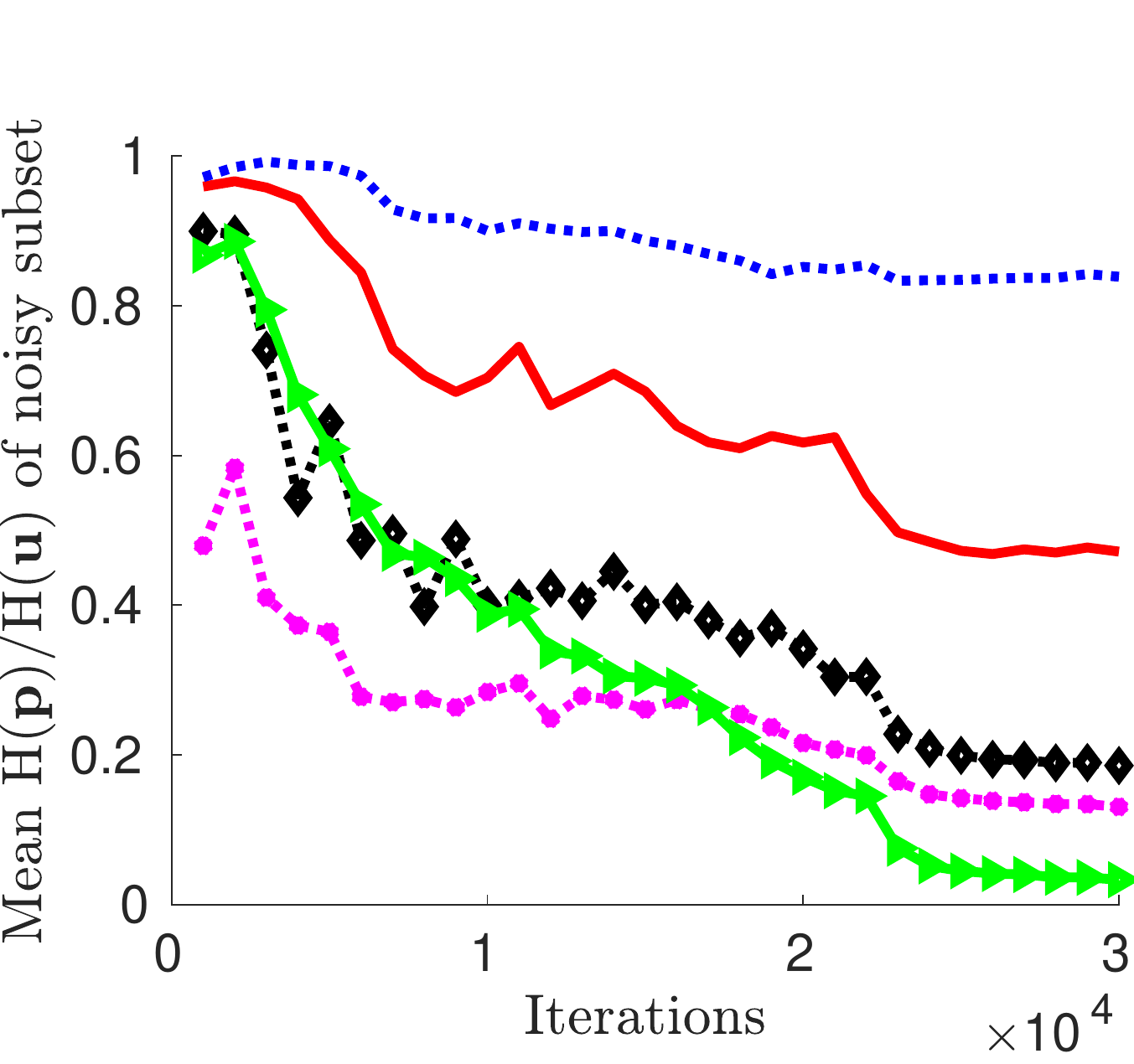}
		\caption{Entropy of noisy subset.}
		\label{fig:CIFAR100Asy0_4_mean_entropy_NoisySubset}
	\end{subfigure}
	\begin{subfigure}[!h]{0.33\textwidth}
		\centering
		\captionsetup{width=\textwidth}
		\includegraphics[clip, trim=3.05cm 8.9cm 4.2cm 7.9cm, width=0.99\textwidth]{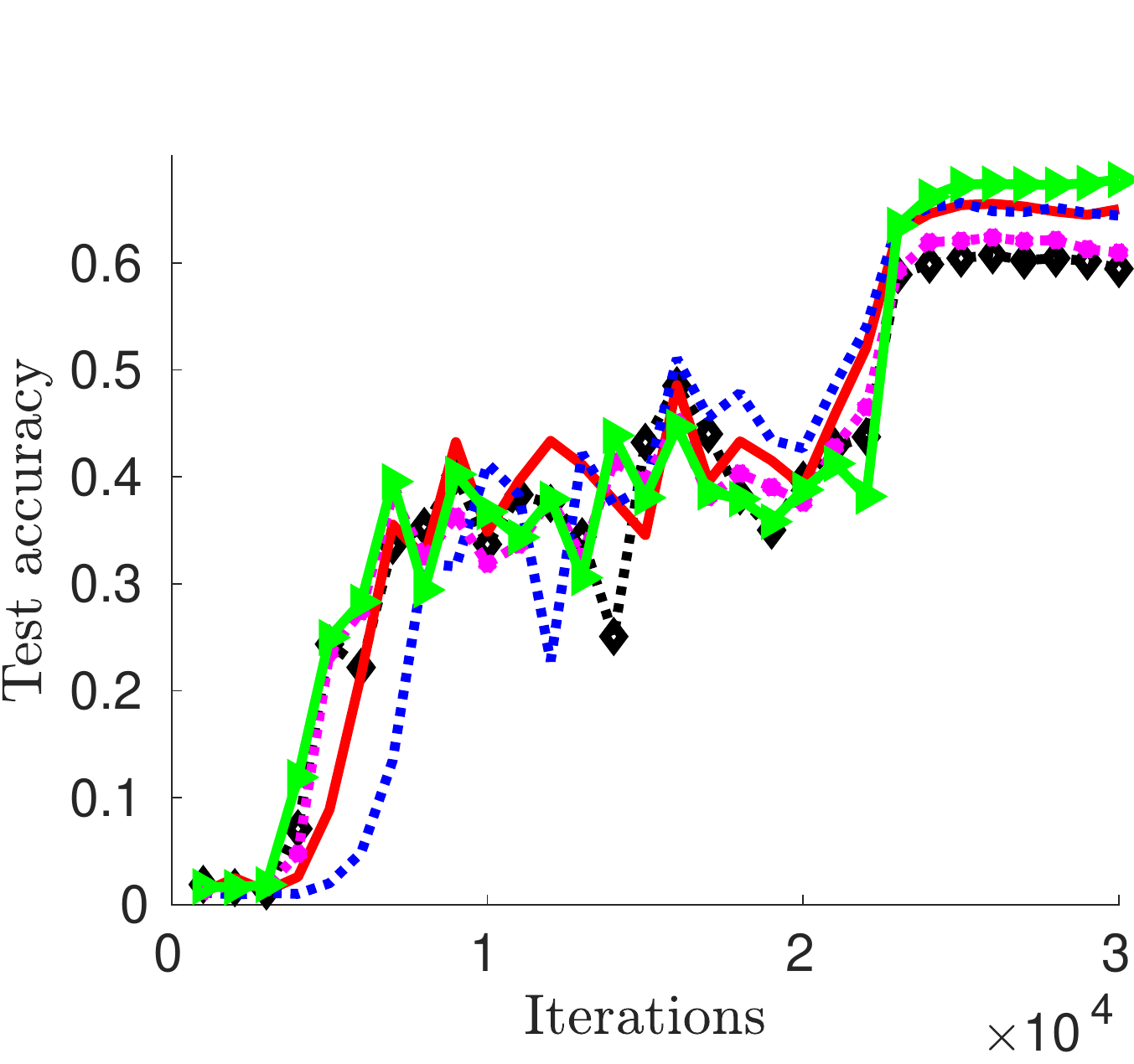}
		\caption{Generalisation.}
		\label{fig:CIFAR100Asy0_4_accuracy30000}
	\end{subfigure}

	\vspace{-0.22cm}
	\caption{ 
		Comprehensive learning dynamics on CIFAR-100 with $40\%$ asymmetric label noise. 
		Vertical axes denote evaluation metrics.    
		Their mean results are displayed.  
		{\textit{At training, a learner is NOT GIVEN whether a label is trusted or not. }}
		We store intermediate models and analyse them when the training ends. 
	}
	\label{fig:comprehensive_dynamics}
	\vspace{-0.38cm}
\end{figure*}

(2)
{{Revise the semantic class when $t$ is large enough and $\mathbf{p}$ is confidently inconsistent.}}
As highlighted in Table~\ref{table:case_analysis_ProSelfLC}, when two conditions are met, we have $\epsilon_{\mathrm{ProSelfLC}}  > 0.5$ and 
$\argmax\nolimits_j \mathbf{p}(j|\mathbf{x}) \neq \argmax\nolimits_j \mathbf{q}(j|\mathbf{x})$, then $\mathbf{p}$ redefines the semantic class. 
For example, if $\mathbf{p} = [0.95, 0.01, 0.04], \mathbf{q} = [0, 0, 1], \epsilon_{\mathrm{ProSelfLC}}=0.8 \Rightarrow \mathbf{\tilde{q}_{\mathrm{ProSelfLC}}}=(1- \epsilon_{\mathrm{ProSelfLC}})  \mathbf{q}+\epsilon_{\mathrm{ProSelfLC}} \mathbf{p}=[0.76, 0.008, 0.232]$.  
%
Note that ProSelfLC also becomes robust against 
lengthy exposure
to the training data, as demonstrated in Figures~\ref{fig:AblationNoEntropy}, \ref{fig:comprehensive_dynamics}.  

\vspace{-0.1cm}
\section{Experiments}
\vspace{-0.1cm}

In deep learning, small differences (e.g., random accelerators like cudnn and different frameworks like Caffe \cite{jia2014caffe}, Tensorflow \cite{abadi2016tensorflow} and PyTorch \cite{paszke2019pytorch}) may lead to a large gap of final performance. 
Therefore, to compare more properly, we re-implement CCE, LS and CP. Regarding Self LC methods, we re-implement Boot-soft \cite{reed2015training}, where $\epsilon$ is fixed throughout training.  
We do not re-implement stage-wise Self LC and KD methods, e.g., Joint Optimisation and Tf-KD$_{self}$ respectively, because time-consuming tuning is required. 
%
%
We fix the random seed and do not use any random accelerator.
%
%
In standard and synthetic cases, 
we train on 80\% training data (corrupted in synthetic cases) and use 20\% trusted training data as a validation set to search all hyperparameters, e.g., $\epsilon, \Gamma,\emB$ and settings of an optimiser.  
\textit{Note that $\Gamma$ and an optimiser's settings are searched first and then fixed for all methods.}  
Finally, we retrain a model on the entire training data (corrupted in synthetic cases) and report its accuracy on the test data to fairly compare with prior results. 
In real-world label noise, the used dataset has a separate clean validation set.
%
Here, a clean dataset is used only for validation, which is generally necessary for any method 
and differs from the methods \cite{vahdat2017toward,veit2017learning,jiang2018mentornet,ren2018learning,li2017learning,hendrycks2018using,yeung2017learning,zhang2020distilling} which use a clean dataset to train a network's learning parameters. 
%

\begin{table*}[!t]
	\vspace{-0.10cm}
	\caption{
		{
			{
				Test accuracy (\%) in the standard setting. 
				We report three settings of hyperparameters.
		}}
	}
	\centering
	\fontsize{9.5pt}{9.5pt}\selectfont
	
	\vspace{-0.27cm}
	\begin{tabular}{lccccccccccccc}
		\toprule
		\multirow{2}{*}{Dataset}
		&\multirow{2}{*}{CCE} & 
		\multicolumn{3}{c}{LS ($\epsilon$)} &   \multicolumn{3}{c}{CP ($\epsilon$)} &  
		\multicolumn{3}{c}{Boot-soft ($\epsilon$)} &  
		\multicolumn{3}{c}{ProSelfLC ($\emB$)}
		\\
		\cmidrule(lr){3-5}
		\cmidrule(lr){6-8}
		\cmidrule(lr){9-11}
		\cmidrule(lr){12-14}
		&
		& $0.125$ & $0.25$ & $0.50$ 
		& $0.125$ & $0.25$ & $0.50$ 
		& $0.125$ & $0.25$ & $0.50$ 
		& $8$ & $10$ & $12$ 
		\\
		\midrule
		CIFAR-100 &69.0 & 69.9 & 69.6 & 68.4 & 69.5 & 69.3 & 68.7 & 68.9 & 69.1 & 69.1 & {70.1} & \textbf{70.3} & {69.8}\\
		ImageNet 2012 &75.5 & 75.3 & 75.2 & 74.9 & 75.2 & 74.8 & 74.6 & 75.7 & 75.8 & 75.8 & \textbf{76.0} & \textbf{76.0} & {75.9}\\
		\bottomrule
	\end{tabular}
	\label{table:ImageNet2012}
	\vspace{-0.35cm}
\end{table*}

\vspace{-0.1cm}
\subsection{Standard image classification}
\label{sec:imagenet2012}
\vspace{-0.10cm}


%
\textbf{Datasets and training details.}
(1) {CIFAR-100} \cite{krizhevsky2009learning} has 20 coarse classes, each containing 5 fine classes. 
There are 500 and 100 images per class in the training and testing sets, respectively. 
The image size is $32 \times 32$. 
We apply simple data augmentation \cite{he2016deep}, i.e., we pad 4 pixels on every side of the image, and then randomly crop it with a size of $32\times 32$. 
Finally, this crop is horizontally flipped with a probability of 0.5.
We choose SGD with its settings as: (a) a learning rate of 0.1; (b) a momentum of 0.9; (c) a weight decay of $5e-4$; (d) the batch size is 256 and the number of training iterations is 30k. 
We divide the learning rate by 10 at 15k and 22k iterations, respectively.  
(2) We train ResNet-50 \cite{he2016deep} on ImageNet 2012 classification dataset, which has 1k classes and 50k images in the test set \cite{russakovsky2015imagenet}.  
%
We use SGD with a start learning rate of $2e-3$. A polynomial learning rate decay with a power of 2 is used. 
We set the momentum to 0.95 and the weight decay to $1e-4$. 
We train on a single V100 GPU and the batch size is 64. 
We report the final test accuracy when the training ends at 500k iterations. 
We use the standard data augmentation: an original image is warped to $256\times256$, followed by a random crop of $224\times224$. 
This crop is randomly flipped. 
%
We fix common settings to fairly compare CCE, LS, CP, Boot-soft and ProSelfLC.

\textbf{Result analysis.} 
In Table~\ref{table:ImageNet2012}, we observe the superiority of ProSelfLC in standard setting without considering label noise. 
Being probably surprising, LS and CP reduce the performance consistently as $\epsilon$ increases on ImageNet. Instead, Boot-soft and ProSelfLC improve versus CCE. We remark that both test sets are large so that their differences are noticeable. 


\begin{table*}
	\vspace{-0.10cm}
	\captionsetup{width=\textwidth}
	\caption{
		Accuracy (\%) on the CIFAR-100 clean test set. 
		All compared methods use ResNet-44. 
	}
	\label{table:cifar100_SOTA_Asymmetric_Symmetric}
	\centering
	\fontsize{9.5pt}{9.5pt}\selectfont
	\setlength{\tabcolsep}{7.2pt} 
	
	\vspace{-0.28cm}
	\begin{tabular}{ccccc c ccc}
		\toprule
		\multirow{2}{*}{} & 
		& \multicolumn{3}{c}{Asymmetric Noisy Labels} &~~~& \multicolumn{3}{c}{Symmetric Noisy Labels} \\
		\cmidrule(lr){3-5} \cmidrule(lr){7-9}
		& Method & ~~~$r$=0.2 & ~~~$r$=0.3 & $r$=0.4 & ~~~ & ~~~$r$=0.2 & ~~~$r$=0.4 & $r$=0.6\\
		\midrule
		
		\multirow{5}{*}{\specialcell{Results From \\ SL \cite{wang2019symmetric}}}
		&Boot-hard & ~~~63.4 & ~~~63.2 & 62.1 & & ~~~57.9 & ~~~48.2 & 12.3 \\
		
		&Forward & ~~~64.1 & ~~~64.0 & 60.9 && ~~~59.8 & ~~~53.1 & 24.7 \\
		&D2L & ~~~62.4 & ~~~63.2 & 61.4 &&
		~~~59.2 & ~~~52.0 & 35.3 \\
		
		&GCE & ~~~63.0 & ~~~63.2 & 61.7 &&
		~~~59.1 & ~~~53.3 & 36.2 \\
		&SL & ~~~{65.6} &  ~~~{65.1} & {63.1} &&
		~~~{60.0} &  ~~~{53.7} &  {41.5} \\
		
		\midrule
		\multirow{5}{*}{\specialcell{Our Trained \\Results}}
		
		&CCE & ~~~{66.6} & ~~~{63.4} & {59.5}&&
		~~~{58.0} & ~~~{50.1} & {37.9} \\
		
		&LS & ~~~{67.9} & ~~~{66.4} & {65.0}&&
		~~~{63.8} & ~~~{57.2} & {46.5} \\
		
		&CP & ~~~{67.7} & ~~~{66.0} & {64.4}&&
		~~~{64.0} & ~~~{56.8} & {44.1} \\
		
		&Boot-soft & ~~~{66.9} & ~~~{65.3} & {61.0}&& 
		~~~{63.2} & ~~~{59.0} & {44.8} \\
		
		
		
		&ProSelfLC  & ~~~\textbf{68.7} & ~~~\textbf{68.5} & \textbf{67.9} &&
		~~~\textbf{64.8} & ~~~\textbf{59.3} & \textbf{47.7} \\
		
		\bottomrule
	\end{tabular}
	\vspace{-0.08cm}
\end{table*}

\begin{table*}[!t]
	\caption{
		{
			{
				The results of different hyperparameters on CIFAR-100 using ResNet-44. 
				Under different noise rates, the best instantiation of each approach is bolded except for CCE. 
		}}
	}
	\centering
	\fontsize{9.5pt}{9.5pt}\selectfont
	\vspace{-0.28cm}
	\begin{tabular}{ccccc|ccc|c}
		\toprule
		\multirow{2}{*}{\makecell{Method \\(hyperparameter)}} & \multirow{2}{*}{\makecell{Value of \\hyperparameter}} & 
		\multicolumn{3}{c}{\makecell{Asymmetric label noise }}
		& \multicolumn{3}{c}{\makecell{Symmetric label noise }} 
		& \multicolumn{1}{c}{\multirow{2}{*}{\makecell{Clean}}}\\
		&&~~20\%&~~30\%&40\%&~~20\%&~~40\%&60\%\\
		\hline
		
		~\\
		CCE & None or $\epsilon=0$ & 66.6 & 63.4 & 59.5 & 58.0 & 50.1 & 37.9 & 69.0\\
		\cmidrule{2-9}
		
		\multirow{3}{*}{LS ($\epsilon$)}&0.125 & 66.4 & 65.6  & 63.1 & 61.7 & 52.5 & 39.1 & \textbf{69.9} \\
		&0.25 & \textbf{67.9} & \textbf{66.4} & \textbf{65.0} & 62.8 & 55.9 & 40.9 & 69.6\\
		&0.50 & 66.8 & 65.8 & 64.6 & \textbf{63.8} & \textbf{57.2} & \textbf{46.5} & 68.4\\
		
		\cmidrule{2-9}
		\multirow{3}{*}{CP ($\epsilon$)}&0.125 & 65.7 & 64.2 & 60.3 & 59.8 & 52.3 & 39.6 & \textbf{69.5} \\
		&0.25 & 66.8 & 65.1 & 61.6  & 61.0 & 53.3 & 40.9 & 69.3\\
		&0.50 & \textbf{67.7} & \textbf{66.0} & \textbf{64.4} & \textbf{64.0} & \textbf{56.8} & \textbf{44.1} & 68.7  \\
		
		\cmidrule{2-9}
		\multirow{3}{*}{Boot-soft ($\epsilon$)}&0.125 & 65.8 & 64.1 & 60.7  & 59.7 & 51.2 & 40.6 & 68.9 \\
		&0.25 & 66.2 & 64.1 & 60.3 & 61.1 & 54.4 & 43.3 & \textbf{69.1} \\
		&0.50 & \textbf{66.9} & \textbf{65.3} & \textbf{61.0} & \textbf{63.2} & \textbf{59.0} & \textbf{44.8} & \textbf{69.1} \\
		
		\cmidrule{2-9}
		\multirow{5}{*}{ProSelfLC ($\emB$)}&8 & 67.8 & 67.4 & \textbf{67.9} & 64.7 & 57.7 & \textbf{47.7} & 70.1\\
		&10 & 68.5 & \textbf{68.5} & 66.8 & 63.9 & 59.0 & 47.5 & \textbf{70.3} \\
		&12 & 68.6 & 67.9 & 67.4 & 64.0 & \textbf{59.3} & 47.5 & 69.8\\
		&14 & \textbf{68.7} & 68.0 & 67.8 & \textbf{64.8} & 59.0 & 47.4 & 69.6  \\
		&16 & 68.4 & 67.2 & 67.3 & 63.7 & 59.0 & 32.3 & 69.9 \\

		\bottomrule
	\end{tabular}
	\label{table:AblationStudyonB1}
	\vspace{-0.32cm}
\end{table*}


\vspace{-0.1cm}
\subsection{Synthetic label noise}
\label{sec:cifar100} 
\vspace{-0.10cm}

\textbf{Noise generation}.  
(1) {Symmetric label noise}: the original label of an image is uniformly changed to one of the other classes with a probability of $r$;
(2) {Asymmetric label noise}: we follow \cite{wang2019symmetric} to generate asymmetric label noise to fairly compare with their reported results. Within each coarse class, we randomly select two fine classes $A$ and $B$. Then we flip $r\times100\%$ labels of $A$ to $B$, and $r\times100\%$ labels of $B$ to $A$. We remark that the overall label noise rate is smaller than $r$. 

\textbf{Baselines.}\footnote{We do not consider DisturbLabel \cite{xie2016disturblabel}, which flips labels randomly and is counter-intuitive. It weakens the generalisation because generally the accuracy drops as the uniform label noise increases.}  
We compare with the results reported recently in SL \cite{wang2019symmetric}.
Forward is a loss correction approach that uses a noise-transition matrix \cite{patrini2017making}.  
D2L monitors the subspace dimensionality change at training \cite{ma2018dimensionality}. 
GCE denotes generalised cross entropy \cite{zhang2018generalized} and SL is symmetric cross entropy \cite{wang2019symmetric}. They are robust losses designed for solving label noise. 
Training details are the same as Section ~\ref{sec:imagenet2012}.
%

\begin{figure*}[!t]
	\vspace{-0.25cm}
	\centering
	\begin{subfigure}[!h]{0.33\textwidth}
		\centering
		\captionsetup{width=\textwidth}
		\vspace{0.52cm}
		\includegraphics[width=1.\textwidth]{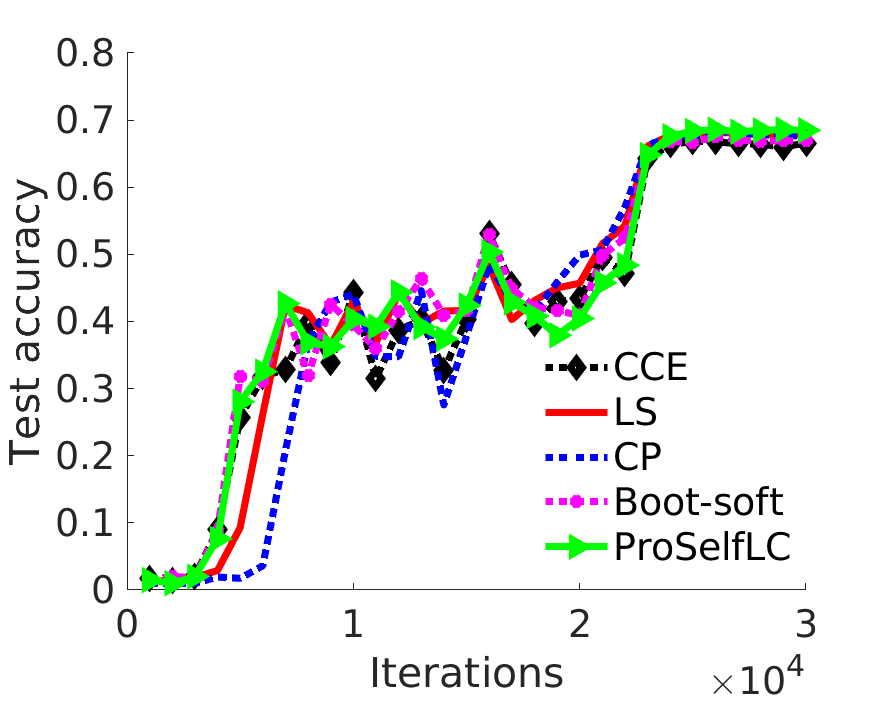}
		\caption{$r=20\%$.}
		\label{}
	\end{subfigure}%
	\begin{subfigure}[t!]{0.33\textwidth}
		\centering
		\captionsetup{width=\textwidth}
		\vspace{0.70cm}
		\includegraphics[width=\textwidth]{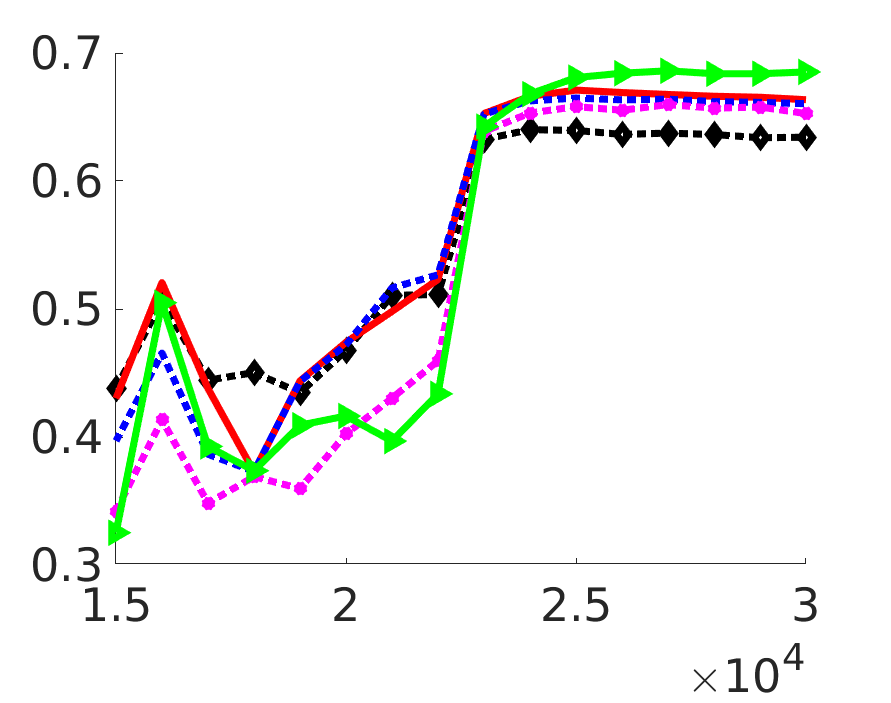}
		\caption{$r=30\%$.}
		\label{}
	\end{subfigure}%
	\begin{subfigure}[t!]{0.33\textwidth}
		\centering
		\captionsetup{width=\textwidth}
		\vspace{0.70cm}
		\includegraphics[width=\textwidth]{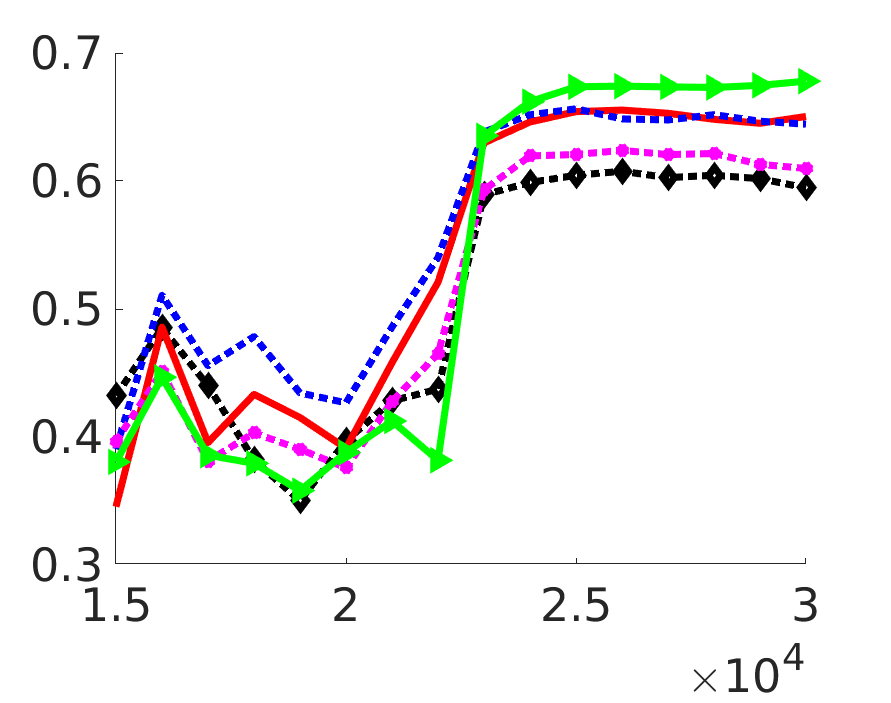}
		\caption{$r=40\%$.}
		\label{}
	\end{subfigure}%
	
	
	\begin{subfigure}[t!]{0.33\textwidth}
		\centering
		\captionsetup{width=\textwidth}
		\vspace{-0.08cm}
		\includegraphics[width=\textwidth]{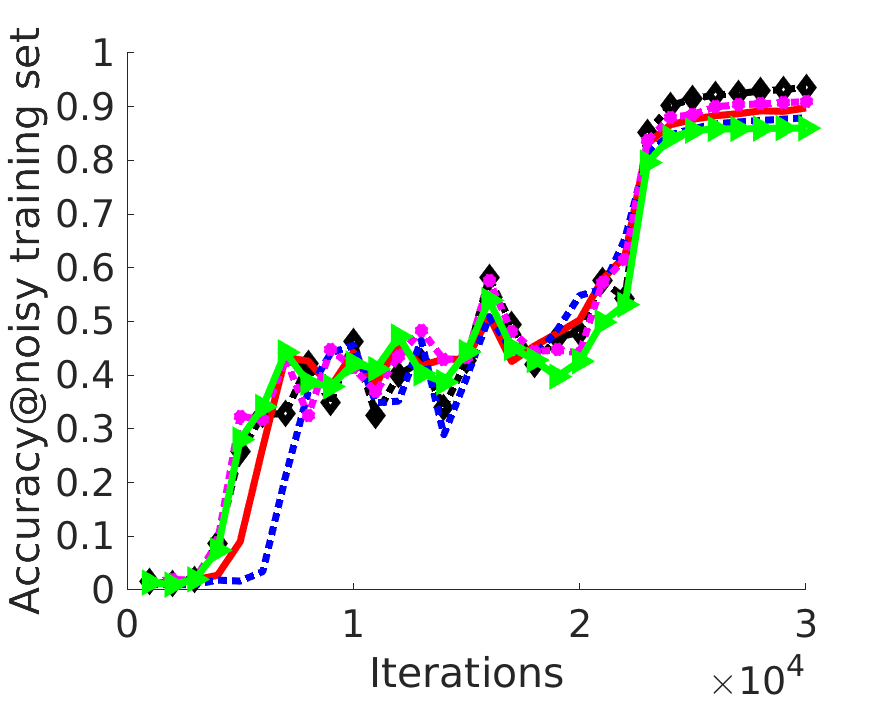}
		\caption{$r=20\%$.}
		\label{}
	\end{subfigure}%
	\begin{subfigure}[t!]{0.33\textwidth}
		\centering
		\captionsetup{width=\textwidth}
		\vspace{0.1cm}
		\includegraphics[width=1.0\textwidth]{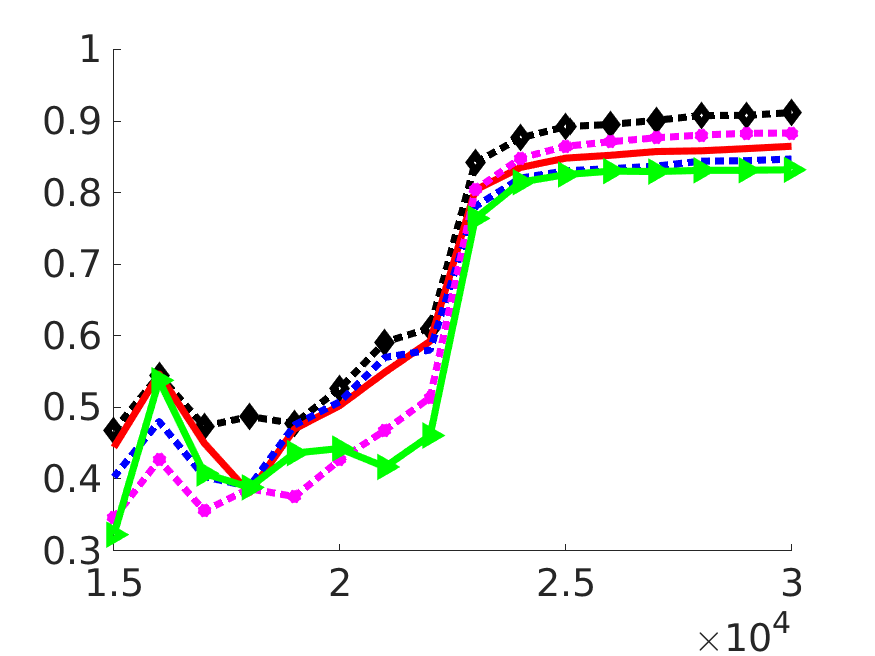}
		\caption{$r=30\%$.}
		\label{}
	\end{subfigure}%
	\begin{subfigure}[t!]{0.33\textwidth}
		\centering
		\captionsetup{width=\textwidth}
		\vspace{-0.1cm}
		\includegraphics[width=1.0\textwidth]{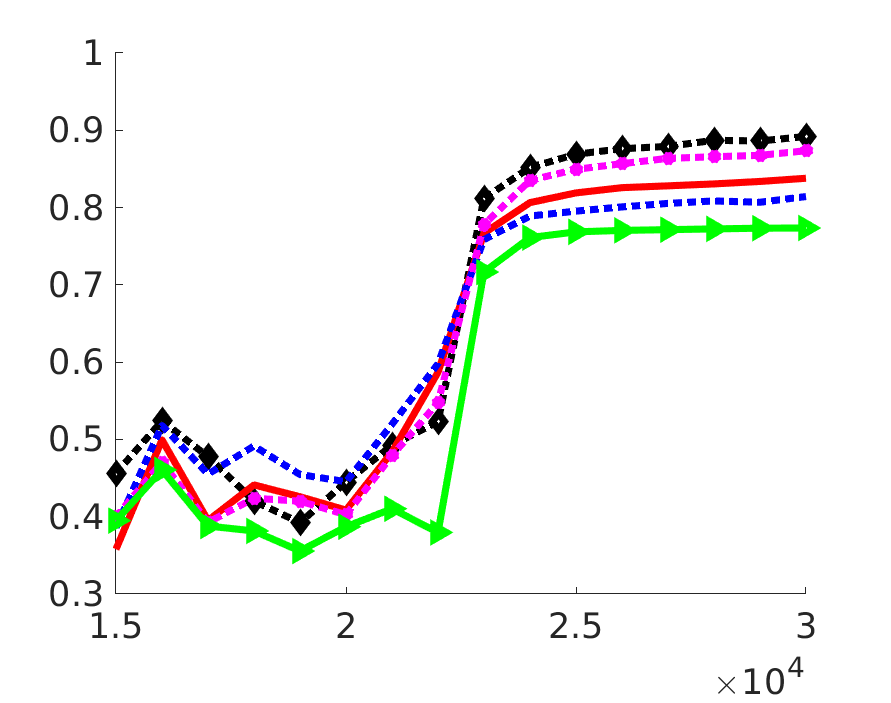}
		\caption{$r=40\%$.}
		\label{}
	\end{subfigure}%
	\vspace{-0.38cm}
	\caption{ 
		Learning dynamics on CIFAR-100 under asymmetric noisy labels. 
		%
		We show all iterations only in (a) and (d). In the others, we show the second half iterations, which are of higher interest. 
		{As the noise rate increases, the superiority of ProSelfLC becomes more significant, i.e., 
			avoiding fitting noise in the 2nd row and 
			better generalisation in the 1st row. }
	}
	\label{fig:learning_dynamics_cifar100_asymmetric}
	\vspace{-0.29cm}
\end{figure*}

\textbf{Result analysis.} 
For all methods, we directly report their final results when training terminates. Therefore, \textit{we test the robustness of a model against not only label noise, but also a long time being exposed to the data.}
In Table ~\ref{table:cifar100_SOTA_Asymmetric_Symmetric}, we observe that: 
(1) 
ProSelfLC outperforms all baselines, which is significant in most cases; 
(2) 
In both implementation, Boot-hard and Boot-soft perform worse than the others. However, our ProSelfLC makes Self LC the best solution.
%
Furthermore, {learning dynamics are visualised in Figure~\ref{fig:comprehensive_dynamics}, which helps to understand why ProSelfLC works better.} 
%
%
%
Figure~\ref{fig:learning_dynamics_cifar100_asymmetric} shows the learning dynamics when $r$ changes. 
Results of different $\emB, \epsilon$ are in Table~\ref{table:AblationStudyonB1}. 
We note that when the noise rate is higher, a smaller $B$ performs better.  

\vspace{-0.08cm}
We further discuss their differences on model calibration \cite{guo2017calibration} in Appendix \ref{appendix:error_and_model_calibration}, the changes of entropy and $\epsilon_{\mathrm{ProSelfLC}}$ during training in Appendix \ref{appendix:entropy_and_coef_dynamics}.


%
%
%
%
%
%
\textbf{Revising the semantic class and similarity structure.} 
In Figures \ref{fig:CIFAR100Asy0_4_corruptedFittting} and ~\ref{fig:CIFAR100Asy0_4_semanticCorrection}, we show dynamic statistics of different approaches on fitting wrong labels and correcting them. ProSelfLC is much better than its counterparts. Semantic class correction reflects the change of similarity structure.  

\begin{table*}[!t]
	\vspace{-0.10cm}
	\caption{
		Test accuracy (\%) on the real-world noisy dataset Clothing 1M. 
	}
	\centering
	\setlength{\tabcolsep}{3.2pt} 
	\fontsize{9.5pt}{9.5pt}\selectfont
	
	\vspace{-0.28cm}
	\begin{tabular}{cccccccccccccc}
		\toprule
		
		\multirow{2}{*}{\makecell{Boot-\\hard}} & \multirow{2}{*}{{Forward} } & 
		\multirow{2}{*}{{D2L}} & 
		\multirow{2}{*}{{~GCE} }  & 
		\multirow{2}{*}{{SL~}} & 
		\multirow{2}{*}{\makecell{~S-\\~adaptation}} &
		\multirow{2}{*}{ Masking~} &
		\multirow{2}{*}{\makecell{~MD-\\DYR-SH}} &
		\multirow{2}{*}{ \makecell{Joint-\\soft~~}} & 
		\multicolumn{5}{c}{Our Trained Results}  
		\\
		\cmidrule{10-14}
		
		&  
		&  &   &  &  &  &  & & CCE & LS & CP & Boot-soft & ProSelfLC\\
		\midrule
		%
		68.9 & 69.8& 69.5 & 69.8 & {71.0}  
		& 70.3 & {71.1} & 71.0 & 72.2 & 
		71.8 &  72.6 & 72.4 & 72.3 & \textbf{73.4} \\
		\bottomrule
	\end{tabular}
	\label{table:Clothing1M_competitors}
	\vspace{-0.69cm}
\end{table*}

\textbf{To redefine and reward a low-entropy status.} 
On the one hand, we observe that LS and CP work well, being consistent with prior claims. 
In Figures~\ref{fig:CIFAR100Asy0_4_mean_entropy_CleanSubset} and \ref{fig:CIFAR100Asy0_4_mean_entropy_NoisySubset}, the entropies of both clean and noisy subsets are much higher in LS and CP, correspondingly their generalisation is the best except for ProSelfLC in Figure~\ref{fig:CIFAR100Asy0_4_accuracy30000}. 
On the other hand, ProSelfLC has the lowest entropy while performs the best, which proves that a learner's confidence does not necessarily weaken its generalisation performance. 
%
%
Instead, {a model needs to be careful with what to be confident in}. 
As shown by Figures \ref{fig:CIFAR100Asy0_4_corruptedFittting} and ~\ref{fig:CIFAR100Asy0_4_semanticCorrection}, ProSelfLC has the least wrong fitting and most semantic class correction, which indicates that a meaningful low-entropy status is redefined.

\vspace{-0.1cm}
\subsection{Real-world label noise}
\vspace{-0.1cm}

\textbf{Clothing 1M} \cite{xiao2015learning} has around 38.46\% label noise in the training data and about 1 million images of 14 classes from shopping websites. Its internal noise structure is agnostic. 

\textbf{Baselines.} 
For loss correction and estimating the noise-transition matrix, S-adaption \cite{goldberger2017training} uses an extra softmax layer, while Masking \cite{han2018masking} exploits human cognition. 
MD-DYR-SH \cite{arazo2019unsupervised} is a combination of three techniques: dynamic mixup (MD), dynamic bootstrapping together with label regularisation (DYR) and soft to hard (SH). 
The other baselines have been introduced heretofore. 

\textbf{Training details.} 
We follow \cite{tanaka2018joint} to train ResNet-50 and initialise it by a trained model on ImageNet. 
%
We follow Section~\ref{sec:imagenet2012} with small changes: the initial learning rate is 0.01 and we train 10k iterations. They are searched on the separate clean validation set.   
%
%

\textbf{Result analysis.} In Table~\ref{table:Clothing1M_competitors}, analogously to CIFAR-100,  we report our trained results of CCE, LS, CP, Boot-soft and ProSelfLC for an entirely fair comparison. 
ProSelfLC has the highest accuracy, which demonstrates its effectiveness again.

\vspace{-0.1cm}
\section{Conclusion}
\vspace{-0.1cm}

We present a thorough mathematical study on several target modification techniques. Through analysis of entropy and KL divergence, we reveal their relationships and limitations. 
To improve and endorse self label correction, we propose ProSelfLC. Extensive experiments prove its superiority over existing methods under standard and noisy settings. 
ProSelfLC enhances the similarity structure information over classes, and rectifies the semantic classes of noisy label distributions.
ProSelfLC is the first approach to trust self knowledge progressively and adaptively. 

ProSelfLC redirects and promotes entropy minimisation, which is in marked contrast to recent practices of confidence penalty \cite{szegedy2016rethinking,pereyra2017regularizing,dubey2018maximum}.

%

%




{\small
	\bibliographystyle{ieee_fullname}
	\bibliography{ICLR2021_ProSelfLC}
}

\newpage
\newpage
\onecolumn
\appendix

\section{Proof of propositions}
\label{appendix_sec:proof_of_propositions}

\textbf{Proposition 3.} \textit{Compared with CCE, LS and CP penalise entropy minimisation while LC reward it.}
%
\\
\textit{Proof.} We can rewrite CCE, LS, CP, and LC from the viewpoint of KL divergence: 
%
\begin{equation}
	\label{eq:KL_CE}
	\begin{aligned}
		L_\mathrm{{CCE}}(\mathbf{{q}}, \mathbf{p}) = \mathrm{H}(\mathbf{q}, \mathbf{p}) 
		& =\mathrm{KL}(\mathbf{q}||\mathbf{p}) + \mathrm{H}(\mathbf{q}, \mathbf{q}) 
		=\mathrm{KL}(\mathbf{q}||\mathbf{p}), 
	\end{aligned}
\end{equation}
{where we have {$\mathrm{H}(\mathbf{q}, \mathbf{q}) = 0$ because $\mathbf{q}$ is a one-hot distribution}}. 
\begin{equation}
	\label{eq:KL_LS}
	\begin{aligned}
		L_\mathrm{{CCE+LS}}(\mathbf{{q}}, \mathbf{p}; \epsilon) 
		&
		= (1-\epsilon) \mathrm{KL}(\mathbf{q}||\mathbf{p})
		+
		\epsilon \mathrm{KL}(\mathbf{u}||\mathbf{p}) 
		+ \epsilon\mathrm{H}(\mathbf{u},\mathbf{u})
		\\
		&= (1-\epsilon) \mathrm{KL}(\mathbf{q}||\mathbf{p})
		+
		\epsilon \mathrm{KL}(\mathbf{u}||\mathbf{p}) 
		+ \epsilon \cdot \text{constant},
	\end{aligned}
\end{equation}
\begin{equation}
	\label{eq:KL_CP}
	\begin{aligned}
		L_\mathrm{{CCE+CP}}(\mathbf{{q}}, \mathbf{p}; \epsilon) 
		&= (1-\epsilon) \mathrm{KL}(\mathbf{q}||\mathbf{p})
		-
		\epsilon (\mathrm{H}(\mathbf{p},\mathbf{u}) 
		- \mathrm{KL}(\mathbf{p}||\mathbf{u})
		)\\
		& = (1-\epsilon) \mathrm{KL}(\mathbf{q}||\mathbf{p})
		+
		\epsilon \mathrm{KL}(\mathbf{p}||\mathbf{u})
		- 
		\epsilon \cdot \text{constant},
	\end{aligned}
\end{equation}
where $\mathrm{H}(\mathbf{p},\mathbf{u})=\mathrm{H}(\mathbf{u},\mathbf{u}) = \text{constant}$.
Analogously, 
LC  in Eq~(\ref{eq:label_correction}) can also be rewritten: 
\begin{equation}
	\label{eq:KL_LC}
	\begin{aligned}
		L_\mathrm{{CCE+LC}}(\mathbf{{q}}, \mathbf{p}; \epsilon) 
		& = (1-\epsilon) \mathrm{KL}(\mathbf{q}||\mathbf{p})
		-
		\epsilon \mathrm{KL}(\mathbf{p}||\mathbf{u})
		+ 
		\epsilon \cdot \text{constant}.
	\end{aligned}
\end{equation}
In LS and CP, both $+\mathrm{KL}(\mathbf{u}||\mathbf{p})$ and $+\mathrm{KL}(\mathbf{p}||\mathbf{u})$ pulls $\mathbf{p}$ towards $\mathbf{u}$. While in LC, the term   $-\mathrm{KL}(\mathbf{p}||\mathbf{u})$ pushes $\mathbf{p}$ away from $\mathbf{u}$.   \hfill\(\Box\)

\textbf{Proposition 4.} \textit{In CCE, LS and CP, a data point $\mathbf{x}$ has the same semantic class. In addition, $\mathbf{x}$ has an identical probability of belonging to other classes except for its semantic class.
}
\\
\textit{Proof.}  
In LS, the target is $ \mathbf{\tilde{q}_{\mathrm{LS}}}=(1-\epsilon)\mathbf{q}+\epsilon \mathbf{u}$. 
For any $0 \leq \epsilon < 1$, the semantic class is not changed, because $1-\epsilon + \epsilon * \frac{1}{C} > \epsilon * \frac{1}{C}$. 
In addition, $j_1 \neq y, j_2 \neq y \Rightarrow \mathbf{\tilde{q}_{\mathrm{LS}}}(j_1|\mathbf{x}) = \mathbf{\tilde{q}_{\mathrm{LS}}}(j_2|\mathbf{x})=\frac{\epsilon}{C}$.

In CP, $ \mathbf{\tilde{q}_{\mathrm{CP}}}=(1-\epsilon)\mathbf{q}-\epsilon \mathbf{p}$.
In terms of label definition, \textit{CP is against intuition because these zero-value positions in $\mathbf{q}$ are filled with negative values in $\mathbf{\tilde{q}_{\mathrm{CP}}}$.}  
A probability has to be not smaller than zero. So we rephrase $\mathbf{\tilde{q}_{\mathrm{CP}}}(y|\mathbf{x})=(1-\epsilon)-\epsilon*\mathbf{p}(y|\mathbf{x})$, and $\forall j\neq y, \mathbf{\tilde{q}_{\mathrm{CP}}}(j|\mathbf{x})=0$ by replacing negative values with zeros, as illustrated in Figure \ref{fig:LS_CP}.     \hfill\(\Box\)

\section{Discussions on wrongly confident predictions and model calibration}
\label{appendix:error_and_model_calibration}

1. \textit{It is likely that some highly confident predictions are wrong. Will ProSelfLC suffer from an amplification of those errors? }
	
First of all, ProSelfLC alleviates this issue a lot and makes a model confident in correct predictions, according to Figure~\ref{fig:CIFAR100Asy0_4_mean_entropy_NoisySubset} together with \ref{fig:CIFAR100Asy0_4_corruptedFittting} and \ref{fig:CIFAR100Asy0_4_semanticCorrection}. 
\textit{\textbf{Figure~\ref{fig:CIFAR100Asy0_4_mean_entropy_NoisySubset} shows the confidence of predictions, whose majority are correct according to Figure~\ref{fig:CIFAR100Asy0_4_corruptedFittting} and \ref{fig:CIFAR100Asy0_4_semanticCorrection}.}} 
In Figure~\ref{fig:CIFAR100Asy0_4_corruptedFittting}, ProSelfLC fits noisy labels least, i.e., around 12\% so that the correction rate of noisy labels is about 88\% in Figure~\ref{fig:CIFAR100Asy0_4_semanticCorrection}.
Nonetheless, ProSelfLC is non-perfect. A few noisy labels are memorised with high confidence. \\

2. \textit{How about the results of model calibration using a computational evaluation metric: Expected Calibration Error (ECE) \cite{naeini2015obtaining,guo2017calibration}?}

Following the practice of \cite{guo2017calibration}, on the CIFAR-100 test set, we report the ECE (\%, \#bins=10)  of 
ProSelfLC versus CCE, as a complement of Figure ~\ref{fig:comprehensive_dynamics}. 
For a comparison, CCE's results are shown in corresponding brackets.   
We try several confidence metrics (CMs), including probability, entropy, and their temperature-scaled variants using a parameter $T$.
Though the ECE metric is sensitive to CM and $T$, ProSelfLC's ECEs are smaller than CCE's. 

\begin{table}[!h]
	\centering
	\caption{
		ECE results of multiple combinations of logits scaling (logits$/ T $) and confidence metrics (probability and entropy). 
	}
	\setlength{\tabcolsep}{6pt} 
	\vspace{-0.35cm}
	\begin{tabular}{lccccc}
		\midrule
		\makecell{Scaling logits with a temperature parameter $T$:  \\logits$/ T $} & $T =1$ & $T =1/4$ & $T =1/8$ \\
		
		\midrule
		CM =  $\max\nolimits_j \mathbf{p}(j|\mathbf{x})$& 
		15.71 (40.98) & 4.24 (18.27) & \textbf{2.39} (9.94)
		\\
		
		CM = 
		$1 - 
		{\mathrm{H}(\mathbf{p})}/{\mathrm{H}(\mathbf{u})}$&
		17.38 (42.83) & 5.22 (17.84) & \textbf{2.66} (9.53)
		\\
		\midrule
	\end{tabular}
	\label{table:ddd}
\end{table}

\section{The changes of entropy statistics and $\epsilon_{\mathrm{ProSelfLC}}$  at training}
\label{appendix:entropy_and_coef_dynamics}

\begin{figure*}[!h]
	\centering
	\begin{subfigure}{0.49\linewidth}
		\captionsetup{width=0.9\textwidth}
		\includegraphics[clip, trim=1.24cm 7.5cm 1.7cm 5.9cm, width=0.98\textwidth]{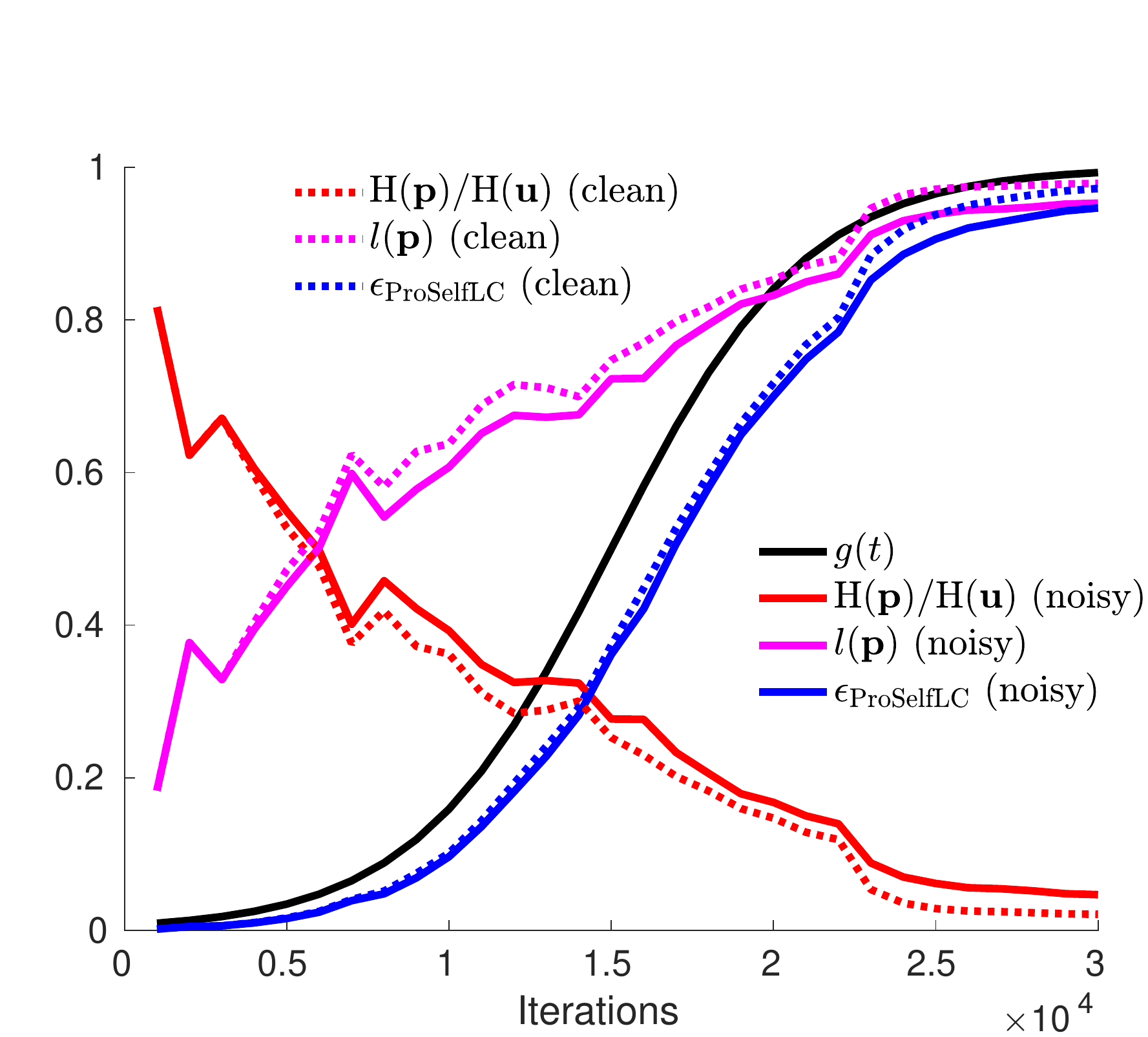}
		\caption{ Asymmetric label noise rate = 20\%.  }
		\label{fig:NoisyCleanSubsets_lcs10_0_2_entropy_episolon}
	\end{subfigure}
	\begin{subfigure}{0.49\linewidth}
		\captionsetup{width=0.9\textwidth}
		\includegraphics[clip, trim=1.24cm 7.5cm 2.3cm 5.9cm, width=0.98\textwidth]{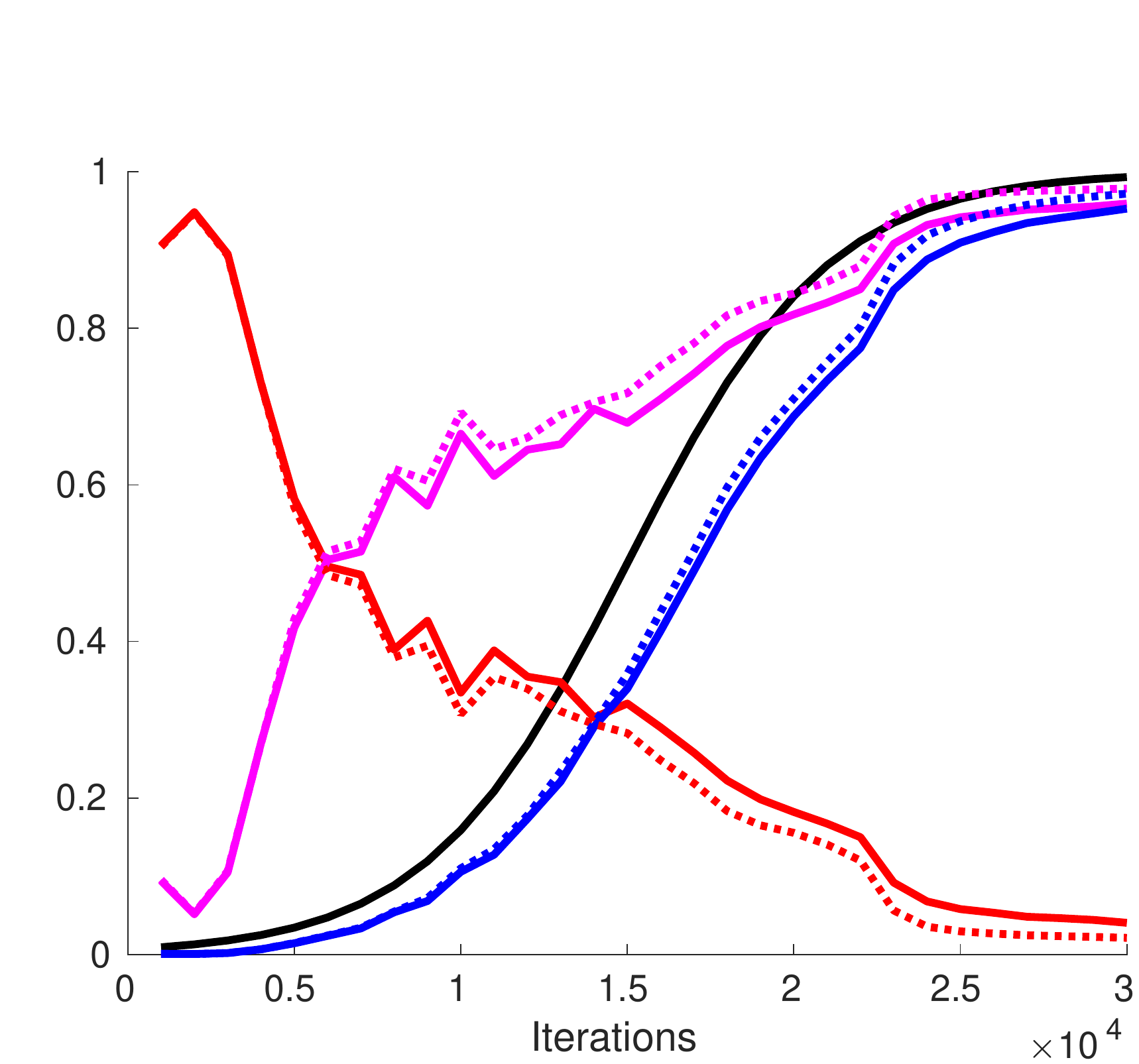}
		\caption{ Asymmetric label noise rate = 40\%. }
		\label{fig:NoisyCleanSubsets_lcs10_0_6_entropy_episolon}
	\end{subfigure}
	\caption{The changes of entropy statistics and $\epsilon_{\mathrm{ProSelfLC}}$  at training.
		We store a model every 1000 iterations to monitor the learning process. 
		For data-dependent metrics, after training, we split the corrupted training data into clean and noisy subsets according to the information about how the training data is corrupted before training.  
		Finally, we report the mean results of each subset.  
	}
	\label{fig:entropy_episoln_dynamics}
\end{figure*}

In Figure~\ref{fig:entropy_episoln_dynamics}, we visualise how the entropies of noisy and clean subsets change at training. 

\end{document}